\documentclass{article}

\usepackage{microtype}
\usepackage{graphicx}
\usepackage{subcaption}
\usepackage{booktabs} % for professional tables\
\usepackage{pifont}
\newcommand{\cmark}{\ding{51}}
\newcommand{\xmark}{\ding{55}}
\usepackage{hyperref}

\usepackage[preprint]{icml2026}

\usepackage{amsmath}
\usepackage{amssymb}
\usepackage{mathtools}
\usepackage{amsthm}
\usepackage{xcolor} % for blue color in new additions
\usepackage[capitalize,noabbrev]{cleveref}

\theoremstyle{plain}
\newtheorem{theorem}{Theorem}[section]

\newtheorem{lemma}[theorem]{Lemma}

\theoremstyle{definition}

\theoremstyle{remark}

\newtheorem*{proposition*}{Proposition}

\newcommand{\vw}{\mathbf{w}}
\newcommand{\vz}{\mathbf{z}}
\newcommand{\vx}{\mathbf{x}}

\newcommand{\f}{f(\vx)}

\newcommand{\nablaf}{\nabla f(\vx)}
\newcommand{\nablafnorm}{\|\nablaf\|}

\newcommand{\cholesky}{\mathbf{L}}
\newcommand{\munabla}{\mathbf{\mu^\nabla(x)}}
\newcommand{\Sigmanabla}{\mathbf{\Sigma^\nabla(x)}}
\newcommand{\vphi}{\boldsymbol{\phi}}
\newcommand{\vPhi}{\boldsymbol{\Phi}}

\usepackage[textsize=tiny]{todonotes}

\icmltitlerunning{Beyond Objective-Based Improvement: Stationarity-Aware Expected Improvement for Bayesian Optimization}

\begin{document}

\twocolumn[
  \icmltitle{Beyond Objective-Based Improvement: Stationarity-Aware Expected Improvement for Bayesian Optimization}



  \icmlsetsymbol{equal}{*}

  \begin{icmlauthorlist}
    \icmlauthor{Joshua Hang Sai Ip}{yyy}
    \icmlauthor{Georgios Makrygiorgos}{yyy}
    \icmlauthor{Ali Mesbah}{yyy}
  \end{icmlauthorlist}

  \icmlaffiliation{yyy}{Department of Chemical and Biomolecular Engineering, University of California, Berkeley, CA, USA}

  \icmlcorrespondingauthor{Joshua Hang Sai Ip}{ipjoshua@berkeley.edu}

  \icmlkeywords{Bayesian Optimization}

  \vskip 0.3in
]



\printAffiliationsAndNotice{}  % no special notice (required even if empty)

\begin{abstract}
Bayesian Optimization (BO) is a principled framework for optimizing expensive black-box functions, with Expected Improvement (EI) among its most widely used acquisition functions. Despite its empirical success, EI is agnostic to first-order optimality conditions, relying solely on objective-value improvement. As a result, it can exhibit vanishing acquisition signals where the improvement criterion is uninformative, limiting its effectiveness in guiding search. We propose Expected Improvement via Gradient Norms (EI-GN), a novel acquisition function that extends the improvement principle to incorporate first-order stationarity, promoting sampling in regions that are both high-performing and close to stationary points. We derive a tractable closed-form expression for EI-GN and show that it remains consistent with the improvement-based acquisition framework. By embedding progress toward stationarity into the acquisition criterion, EI-GN provides a richer and more informative notion of improvement. Empirical results on standard BO benchmarks demonstrate consistent gains over baseline methods, and we further illustrate its applicability to control policy learning.
\end{abstract}

\section{Introduction}

Bayesian Optimization (BO) has emerged as a powerful framework for optimizing expensive black-box functions, with applications spanning materials discovery \cite{lookman2019active}, robotics \cite{calandra2016bayesian}, and hyperparameter optimization in machine learning \cite{wu2019hyperparameter}. Central to BO are probabilistic surrogate models, which guide the selection of query points by balancing exploration and exploitation through acquisition functions (AFs).

Classical BO is typically performed in a derivative-free setting, assuming that gradient information of the objective is unavailable \cite{shahriari2015taking, frazier2018tutorial}. However, this assumption has been revisited in recent work \cite{wu2017bayesian}, as gradient observations are accessible in many practical settings at negligible additional cost. Examples include policy gradients in stochastic control via the policy gradient theorem \cite{sutton1999policy}, adjoint-based sensitivities in PDE-constrained optimization \cite{jameson1999re, plessix2006review}, and hypergradients in machine learning \cite{maclaurin2015gradient}. However, most of the existing approaches to gradient-enhanced BO focus on incorporating gradient observations into the surrogate model to improve predictive accuracy \cite{padidar2021scaling}. While effective, this strategy can become computationally burdensome due to the need to model cross-correlations between function values and partial derivatives across dimensions. In contrast, comparatively little attention has been given to leveraging gradient information directly within the acquisition function \cite{makrygiorgos2023no}.

\begin{figure}[tb]
  \begin{center}
    \centering
    \includegraphics[width=0.85\columnwidth]{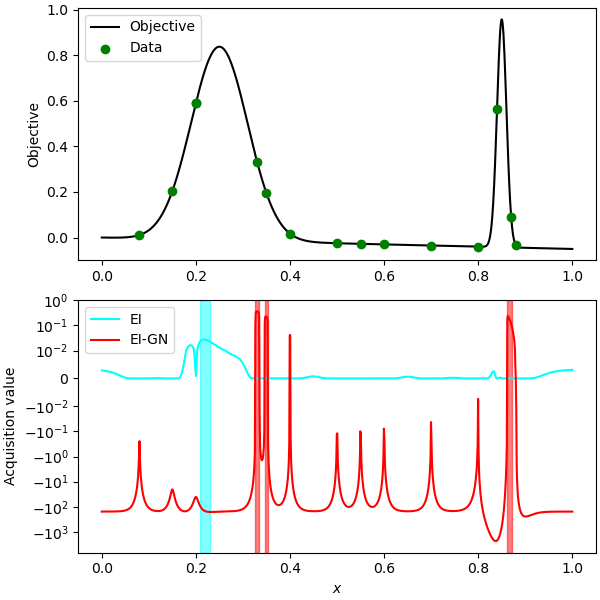}
    \caption{
        Comparison of acquistion behavior for a univariate mixture of Gaussians with a wide local basin and a narrow basin containing the global maximum in $\vx \in [0, 1]$. \textbf{Top:} Objective landscape and sampled data for GP learning. \textbf{Bottom:}
        Acquisition values for EI (cyan) and EI-GN (red) with regions corresponding to large values shaded. % EI-GN is a signed surrogate acquisition (objective-improvement term minus a stationarity penalty) and can take negative values; optimization depends only on relative values.
    }
    \label{fig: acquisition comparison}
  \end{center}
\end{figure}

In this work, we focus on Expected Improvement (EI) \cite{jones1998efficient}, one of the most widely used acquisition functions due to its principled formulation and strong empirical performance. EI relies solely on objective-value improvement, which can lead to vanishing acquisition signal in flat or near-stationary regions, limiting its effectiveness in guiding exploration. Rather than modifying exploration heuristics or rescaling EI, we propose to generalize the notion of improvement itself. Specifically, we introduce an auxiliary objective that incorporates first-order stationarity, enabling improvement to be measured not only in terms of objective value but also in terms of progress toward stationarity. This perspective yields an acquisition function with multiple pathways to improvement: (i) objective value improvement and (ii) reduction in its gradient norm. Consequently, the acquisition landscape remains informative even in regimes where EI provides little signal.

To illustrate this behavior, consider a univariate mixture of Gaussians $f(\vx) = 0.85\exp(-0.5((\vx - 0.25)/0.06)^2) + \exp(-0.5((\vx-0.85)/0.01)^2) - 0.05\vx$, shown in Figure~\ref{fig: acquisition comparison}. The function exhibits a wide basin with a local maximum and a narrow basin containing the global maximum (top plot). Gaussian process (GP) surrogates are trained on identical samples and used to evaluate EI (cyan) and EI-GN (red), with shaded regions indicating high acquisition values (bottom plot). EI is nearly zero over most of the domain and concentrates its mass near the incumbent, favoring the local basin. In contrast, EI-GN identifies multiple high-value regions, reflecting additional improvement pathways induced by the stationarity-aware objective. This richer acquisition structure arises from extending the improvement criterion beyond objective value alone, rather than introducing stronger exploration mechanisms. The contributions of this paper are as follows:
\begin{itemize}
  \item We propose a novel acquisition function, denoted by $\text{EI}_g$, obtained by applying the improvement principle to a stationarity-aware auxiliary objective, thereby extending improvement-based acquisition beyond objective-value criteria. 
  \item We derive a lower bound on $\text{EI}_g$ that decomposes into EI-type terms capturing improvement in both objective value and stationarity. This analysis motivates EI-GN, a tractable AF that instantiates the same principles and admits a closed-form expression under Gaussian posteriors for practical optimization.
\end{itemize}

\section{Background}
\subsection{Problem Formulation}

We seek to maximize an expensive black-box objective $f(\vx)$ 
\begin{equation}
    \mathbf{x}^* \leftarrow \underset{\mathbf{x} \in \mathcal{X}}{\arg\max}\ f(\mathbf{x}),
\end{equation}
where $\vx \in \mathbb{R}^d$. We observe potentially noisy values of the objective $f$ and its gradients $\nabla f$ as $y$ and $\nabla y$, respectively. We assume that gradient observations can be obtained at negligible additional cost compared to evaluating $f$. % and do not charge them to the evaluation budget.

% $f$ and its gradients $\nabla f$ deterministically and denote their observed quantities as $y$ and $\nabla y$ respectively.

% \begin{equation}
% \label{eq: y}
%     y = f(\mathbf{x}) + \epsilon_f, \quad \epsilon_f \sim \mathcal{N}(0, \sigma_f^2),
% \end{equation}
% \begin{equation}
% \label{eq: nabla y}
%     \nabla y = \nabla f(\mathbf{x}) + \boldsymbol{\epsilon}_{\nabla f}, \quad \boldsymbol{\epsilon}_{\nabla f} \sim \mathcal{N}(\mathbf{0}, \Sigma_{\nabla f}).
% \end{equation}
% \subsection{Bayesian Optimization}
% The prinicple behind BO is to optimize an expensive black-box objective in a sample efficient manner; for brevity we will not delve into the specifics here, we refer the reader to \cite{brochu2010tutorial, garnett2023bayesian}. Instead, we focus on the probabilistic surrogates used in BO. 
\subsection{Gaussian Processes}
GPs are typically the surrogate model of choice in BO due to their smoothness and non-parameteric nature, allowing flexibility in modeling objectives under minimal assumptions \cite{williams2006gaussian}. Let $\mathbf{X} = [\mathbf{x}_1,\ldots,\mathbf{x}_N]^\top$, $\mathbf{y} = [y_1,\ldots,y_N]^\top$, and $\nabla\mathbf{y} = [\nabla y_1,\ldots,\nabla y_N]^\top$, where $N$ denotes the number of samples. The posterior mean and variance for the objective can be described as
\begin{equation}
\label{eq: GP_mean}
    \mu(\mathbf{x}) = m(\mathbf{x}) + k(\mathbf{x}, \mathbf{X})K^{-1}(y - m(\mathbf{X})),
\end{equation}
\begin{equation}
\label{eq: GP_var}
    \sigma(\mathbf{x})^2 = k(\mathbf{x}, \mathbf{x}) - k(\mathbf{x}, \mathbf{X})K^{-1}k(\mathbf{X}, \mathbf{x}),
\end{equation}
where $m(\cdot)$ and $k(\cdot, \cdot)$ denote the mean and kernel function, and $K = k(\mathbf{X}, \mathbf{X}) + \sigma_n^2I$.

\subsection{Expected Improvement}
There is a vast body of literature on different AFs, ranging from decision-theoretic approaches such as Knowledge Gradient \cite{wu2016parallel} to information-centric approaches such as Entropy Search \cite{hernandez2016predictive}. We direct our attention to EI \cite{jones1998efficient}, which 
is defined as $\text{EI}(\mathbf{x}) = \mathbb{E}[\max(f(\mathbf{x})-f(\mathbf{x}^+), 0)]$.
% \begin{equation}
% \label{eq: EI}
%     \text{EI}(\mathbf{x}) = \mathbb{E}\left[\max(f(\mathbf{x})-f(\mathbf{x}^+), 0)\right].
% \end{equation}
Since EI only considers domains that lead to improvement over $f(\mathbf{x}^+)$, the AF value can reduce to zero for large subsets of the search space, leading to over-exploitation \cite{qin2017improving, de2021greed}. \citet{hoffman2011portfolio} proposed portfolio allocation with different probabilities of selecting various AFs, \citet{qin2017improving} suggested a stochastic non-greedy selection for EI, and \citet{benjamins2023self} used adaptive weights in EI to balance exploration and exploitation. In contrast, our proposed method directly reshapes the AF landscape by incorporating gradient information.

\section{Related Work}

\subsection{Gradient-enhanced Gaussian Processes}
\label{sec: gradient GP}
There has been work extending GPs to gradient observations \cite{solak2002derivative}, showing gradients are beneficial in BO \cite{shekhar2021significance}. \citet{wu2017bayesian} suggested a joint GP model, where every $f$ and $\nabla f$ is correlated, though this led to prohibitively expensive complexity $\mathcal{O}((d + 1)^3 N^3)$. This motivated approximations that leverage variational inference and inducing points \cite{padidar2021scaling}, as well as gradient-enhanced BNNs \cite{makrygiorgos2025towards}. In this work, gradients are used for acquisition shaping rather than improving the surrogate of $f$.
% The above methods all increase surrogate accuracy of $f$ via gradients, which is beyond the scope of the work. Since the goal is to analyze the role of gradients in providing auxiliary signal in the acquisition landscape, we adopt $d$ parallel, independent gradient-based GPs trained separately from the zeroth-order GP to avoid confounding gradients with surrogate behavior. We emphasize that since cross-covariances are ignored, the posterior on the objective is inferred in the same manner as vanilla BO, highlighting any potential gains are solely due to acquisition behavior. In other words, gradient information does not improve surrogate accuracy of $f$, or its posterior in inference.  

% For $i = 1, ..., d$, the posterior gradient mean $\mu_i^\nabla(\mathbf{x})$ and variance $\sigma^{\nabla}_i(\mathbf{x})^2$ can be expressed similarly
% \begin{equation}
%     \mu_i^\nabla(\mathbf{x}) = m_i(\mathbf{x}) + k_i(\mathbf{x}, \mathbf{X})\mathcal{K}^{-1}_i(\mathbf{X})(\nabla\mathbf{y} - m_i(\mathbf{X})),
% \end{equation}
% \begin{equation}
%     \sigma^{\nabla}_i(\mathbf{x})^2 = k_i(\mathbf{x}, \mathbf{x}) - k_i(\mathbf{x}, \mathbf{X})\mathcal{K}^{-1}_i(\mathbf{X})k_i(\mathbf{X}, \mathbf{x}),
% \end{equation}
% where $m_i(\cdot)$, $k_i(\cdot, \cdot)$, and $\mathcal{K}_i(\mathbf{X})$ now correspond to their respective dimension $i$. As such, $\mu^\nabla := (\mu_1^\nabla, ..., \mu_d^\nabla)^\top$ and $\Sigma^\nabla := \text{diag}\{\sigma_1^{\nabla2}, ..., \sigma_d^{\nabla2}\}$. 

\subsection{Gradient-enhanced Acquisition Functions}
There has been work on gradient-enhanced AFs. \citet{makrygiorgos2023gradient} formulated a multi-objective AF consisting of zeroth-order and first-order terms, where multi-objective optimization leads to a Pareto frontier in the acquisition ensemble. In contrast, we define a single scalar EI-style acquisition on an auxiliary objective that integrates stationarity as a soft penalty, rather than considering tradeoffs between zeroth- and first-order AFs. This maintains the global improvement structure of EI, while providing valuable acquisition shaping for when standard EI becomes near-zero in the search space.

Another line of work considers first-order optimality more explicitly. \citet{penubothula2021novel} directly searched for points that correspond to objective gradients approximately equal to zero, then utilized the ``significance" criterion to evaluate candidate points. \citet{makrygiorgos2023no} enforced constraints akin to first-order optimality conditions directly in the AF for restricting search space. Our work differs from these approaches, as we do not impose constraints nor attempt to locate stationary points; we use gradient information inside the EI framework to guide AF optimization. This preserves EI's improvement-centric behavior, while providing additional acquisition signal for when standard EI over-exploits.
% Therefore, the prior works utilize ensemble AFs under multi-objective optimization, rank candidates that approximately satisfy stationarity, and impose constraints that mimic first-order optimality conditions. Instead, our work introduces soft stationarity penalties into a single auxiliary objective in the EI framework. 

\subsection{Local Bayesian Optimization via Gradient Descent}
% Gradient-enhanced BO largely falls into two categories: Gradient-enhanced surrogates where gradient observations are directly used in model fitting, and local search via gradient descent.
BO is used to facilitate gradient descent by designing AFs that query points that reduce uncertainty in  gradient inference \cite{muller2021local}. This notion has been extended to optimize for gradient descent direction \cite{nguyen2022local}, and gradient descent in multiple objectives \cite{ip2024preference, ip2025user}. These works infer gradients from zeroth-order function observations for performing local optimization via gradient descent. Our method instead uses gradients to directly reshape the AF for gradient-enhanced global optimization.
% However, we argue that these works are orthogonal to the proposed method, since they do not leverage gradient observations and only infer them from zeroth order information. They also use gradients to locally optimize rather than improving global search by altering AF landscapes.

\section{Method}
\subsection{Model}
\label{sec: model}
%As seen in Section~\ref{sec: gradient GP},  there is prior work where gradient observations are utilized to improve surrogate on $f$ via joint modeling of $(f, \nabla f)$ derivative kernels and cross-covariances. 
Our goal is to analyze the role of gradients in providing an auxiliary signal in the acquisition landscape. To avoid confounding acquisition effects with surrogate modeling, we learn a GP for $f$ and $d$ independent GPs based on the objective's partial derivatives. Hence, this allows us to avoid cross-variances between $f$ and $\nabla f$, as well as correlations between individual dimensions in $\nabla f$. Consequently, the surrogate for $f$ is identical to the one used in vanilla BO, ensuring any potential gains in optimization are solely due to acquisition behavior. The choice of uncorrelated models for $f$ and $\nabla f$ reduces complexity to $\mathcal{O}((d+1)N^3)$. If models are trained in parallel, this \textcolor{black}{will reduce wall-clock time} to $\mathcal{O}(N^3)$, independent of $d$. We model each partial gradient using similar equations to \eqref{eq: GP_mean} and \eqref{eq: GP_var}, except each dimension has its mean and kernel functions. We denote the posterior gradient mean and covariance as $\mu^\nabla$ and $\Sigma^\nabla$, respectively. \textcolor{black}{For this work, we focus on the independent models, but we perform additional comparisons with joint GP models \cite{wu2017bayesian} in Sec~\ref{sec: gradient_ablation}.} % Direct comparisons to gradient-enhanced surrogate models are beyond the scope of this work, as they modify the posterior and will confound our focus on acquisition design.

\begin{algorithm}[tb]
\caption{Expected Improvement via Gradient-Norms}
\label{alg: EI-GN}
\begin{algorithmic}[1]
\STATE{\textbf{Input:} Number of iterations $T$, search space $\mathcal{X}$, black-box objective $(f, \nabla f)^\top$, EI-GN penalty weight $\alpha$}
\STATE{\textbf{Output:} Optimal decision variables $\mathbf{x}^*$}
\STATE Initialize dataset $\mathcal{D}_0 \leftarrow (\mathbf{X}_0, \mathbf{y}_0, \nabla\mathbf{y}_0)$ 
\FOR{$t = 1$ to $T$}
    \STATE Fit GP surrogate model on $(\mathbf{X}_t, \mathbf{y}_t)$
    \STATE Fit gradient GP models on $(\mathbf{X}_t, \nabla\mathbf{y}_t)$
    \STATE Compute EI$_f(\vx) = \mathbb{E}[\max(f(\mathbf{x})-f(\mathbf{x}^+), 0)]$
    \STATE Compute $\overline{\text{EI}}_s(\vx)$ from \eqref{eq: EI_s overline}
    % \STATE $\vx_t^+ \leftarrow \underset{\mathbf{x} \in \mathbf{X}_t}{\arg\max}\ g(\vx)$
    \STATE $\mathbf{x}_{t+1} \leftarrow \underset{\mathbf{x} \in \mathcal{X}}{\arg\max}\ \text{EI-GN}(\vx)$ from \eqref{eq: EI-GN}
    % \STATE Evaluate $y_{t+1} = f(\mathbf{x}_{t+1})$
    % \STATE Observe $\nabla y_{t+1} = \nabla f(\mathbf{x}_{t+1})$
    
    \STATE Evaluate $(y_{t+1}, \nabla y_{t+1})^\top$ %according to \eqref{eq: y}, \eqref{eq: nabla y}
    \STATE $\mathcal{D}_{t+1} \leftarrow \mathcal{D}_t \cup \{(\mathbf{x}_{t+1}, y_{t+1}, \nabla y_{t+1})\}$
\ENDFOR
\STATE \textbf{return} $\mathbf{x}^* \leftarrow \underset{(\mathbf{X}, \mathbf{y})\in\mathcal{D}_T}{\arg\max}\ \mathbf{y}$
\end{algorithmic}
\end{algorithm}
\subsection{Expected Improvement on Auxiliary Objective}
\label{sec: EIGN}
Inspired by Sobolev norms, we define an auxiliary function $g(\mathbf{x})$ that augments $\f$ with an additional gradient term $\nablafnorm_2^2$
\begin{equation}
\label{eq: g}
    g(\vx) = \f - \alpha \nablafnorm_2^2,
\end{equation}
where $\alpha \geq 0$ is a hyperparameter that determines the weight of the gradient norm term in $g(\mathbf{x})$. 
% \color{blue}
% \begin{assumption}[Interior maximizer] \label{ass:interior} The objective $f:\mathcal{X}\to\mathbb{R}$ is differentiable and attains a global maximizer $\vx^\star \in \mathrm{int}(\mathcal{X})$ with $\nabla f(\vx^\star)=\mathbf{0}$. \end{assumption}

% \paragraph{Shared optimum.}
% Under Assumption~\ref{ass:interior}, we have $g(\vx^\star)=f(\vx^\star)$ and 
% $\vx^\star \in \arg\max_{\vx} g(\vx)$.
% \color{black}
% To build intuition for how maximizing $g$ can preserve optima of $f$, consider the idealized smooth setting where $f$ attains a global maximizer $\vx^\star \in \mathrm{int}(\mathcal{X})$ such that $\nabla f(\mathbf{x}^\ast) = \mathbf{0}$.\footnote{\textcolor{blue}{We note the interior maximizer condition may not hold in practice, but is a reasonable assumption for many black-box functions.}} This results in $g(\vx^\star)=f(\vx^\star)$, where $\vxstar$ is also a global maximizer of $g$. 

The motivation for \eqref{eq: g} is derived from the first-order necessary conditions for optimality \cite{boyd2004convex}, where interior optima must respect stationarity conditions. Therefore, the gradient term can be considered as a soft penalty for deviations from stationarity, rather than explicit enforcement of Karush-Kuhn-Tucker (KKT) conditions. Unlike hard constraints, the soft penalty implicitly biases search away from regions that are highly non-stationary, facilitating continued \textcolor{black}{search} even when local basins that satisfy first-order optimality conditions are sparse. 
%However, we recognize that this formulation implicitly assumes the existence of interior global maxima; EI-GN may be less effective for problem landscapes with boundary global maxima where stationarity assumptions do not hold. Yet, empirically we observe that the proposed method still has strong performance in some boundary-maxima problems due to the soft penalty (i.e. Holder Table as seen in Figure~\ref{fig1: holder_contour}.
We define EI for $g(\vx)$ and denote it as $\text{EI}_g$, i.e.,
\begin{equation}
\label{eq: EI-GN}
    \text{EI}_g(\mathbf{x}) = \mathbb{E}\left[\max(g(\vx) - g(\vx^+), 0)\right],    
\end{equation}
where $\vx^+$ denotes the decision variables corresponding to the incumbent in $g$.

\subsection{\textcolor{black}{Lower Bound of} $\text{EI}_g$}
Computing \eqref{eq: EI-GN} does not admit a closed-form expression due to the coupling of function values and gradient norms in $g(\vx)$. Therefore, we separate the zeroth-order and first-order terms and apply the Positive Part Inequality (Lemma~\ref{lemma: positive part inequality}) to \textcolor{black}{form the following lower bound}
\begin{equation}
\label{eq: EI-GN lower bound}
    \text{EI}_g(\mathbf{x}) \geq \text{EI}_f(\vx) - \alpha\text{EI}_s(\vx),
\end{equation}
% \begin{equation}
% \label{eq: EI-GN lower bound}
%     \text{EI}_g(\mathbf{x}) \geq \text{EI}_f(\vx) - \alpha\mathbb{E}[\max(\nablafnorm_2^2 - \|\nabla f(\vx^+)\|_2^2, 0)],
% \end{equation}
where EI$_f$ denotes EI defined on the objective $f$; akin to standard EI. EI$_s$ is defined as 
\begin{equation}
\label{eq: EI_s}
\text{EI}_s(\vx) = \mathbb{E}[\max(\nablafnorm_2^2 - \|\nabla f(\vx^+)\|_2^2, 0)],
\end{equation}
which is an EI-style improvement term in stationarity. The \textcolor{black}{lower bound} in \eqref{eq: EI-GN lower bound} enables us to isolate the zeroth-order term from the first-order term \textcolor{black}{and recovers standard} EI when $\alpha = 0$. The derivation of \eqref{eq: EI-GN lower bound} can be found in Appendix~\ref{app: EI-GN}. %\textcolor{blue}{In practice, we do not optimize $\mathrm{EI}_g$ directly due to intractability; instead, we optimize a tractable proxy derived from its lower bound, which leads to the final acquisition EI-GN.}

\subsection{Acquisition Signal in Low-Improvement Regions}
\label{sec:acquisition_signal}

We now explain why $\text{EI}_g$ provides acquisition signal in low-improvement regions, and why this persists when optimizing its \textcolor{black}{lower bound} \eqref{eq: EI-GN lower bound}. Improvement in $g$ reduces to
% \begin{equation}
% \label{eq:g_decomposition}
% g(\vx) - g(\vx^+) = \underbrace{(f(\vx) - f(\vx^+))}_{\text{objective change}} - \alpha\underbrace{(\|\nabla f(\vx)\|_2^2 - \|\nabla f(\vx^+)\|_2^2)}_{\text{stationarity change}}.
% \end{equation}
\begin{equation}
\begin{aligned}
\label{eq:g_decomposition}
g(\vx) - g(\vx^+)
&= \underbrace{\big(f(\vx) - f(\vx^+)\big)}_{\text{objective change}}\\
&-\alpha\underbrace{\big(\|\nabla f(\vx)\|_2^2 - \|\nabla f(\vx^+)\|_2^2\big)}_{\text{stationarity change}} .
\end{aligned}
\end{equation}
This reveals two pathways to improvement: (i) $f(\vx)$ exceeds $f(\vx^+)$, the standard EI mechanism, or (ii) $\|\nabla f(\vx)\|_2^2 < \|\nabla f(\vx^+)\|_2^2$, i.e., the gradient norm is smaller than the incumbent's, contributing positively even if $f(\vx) < f(\vx^+)$. Crucially, pathway (ii) can yield $g(\vx) > g(\vx^+)$ even when $f(\vx) < f(\vx^+)$; acquisitions based only on zeroth-order improvement (such as EI on $f$) only have access to pathway (i). 

\textcolor{black}{To build intuition for when stationarity improvement contributes to progress in $g$}, we define an event, capturing the trade-off between objective loss and gradient reduction 
\begin{equation}
\begin{aligned}
\label{eq:event_def}
\mathcal{E}_{\delta,c}(\vx) &= \big\{\|\nabla f(\vx)\|_2^2 \leq \|\nabla f(\vx^+)\|_2^2 - \delta\big\}\\ 
&\cap \big\{f(\vx) \geq f(\vx^+) - \alpha c \delta\big\},
\end{aligned}
\end{equation}
\textcolor{black}{where $\delta > 0$ is a stationarity margin in units of squared gradient norm, $c \in (0,1)$ is the tolerated fraction of stationarity improvement in exchange with objective decrease, and $\alpha > 0$ is the weighting parameter balancing the objective and gradient terms in $g$.} \textcolor{black}{On this event, improvements in stationarity can offset objective loss, leading to an increase in $g$. This suggests that $\mathrm{EI}_g$ can remain informative even when $\mathbb{P}(f(\vx) > f(\vx^+))$ is small.}

% On this event, stationarity improves by at least $\delta$, while objective drops by at most $\alpha c \delta$, so the stationarity contribution $\alpha \delta$ outweighs the objective loss. Hence, $\mathcal{E}_{\delta,c}(\vx)$ applied to \eqref{eq:g_decomposition} yields $g(\vx)-g(\vx^+) \ge -\alpha c\delta + \alpha\delta = \alpha(1-c)\delta$. \textcolor{blue}{This yields the inequality} $\mathrm{EI}_g(\vx) \geq \alpha(1-c)\delta\,\mathbb{P}(\mathcal{E}_{\delta,c}(\vx))$. \textcolor{blue}{This suggests} $\mathrm{EI}_g$ can remain informative even when $\mathbb{P}(f(\vx) >  f(\vx^+)) \approx 0$. 
% Revisiting the $\text{EI}_g$ lower bound \eqref{eq: EI-GN lower bound}, we define 
% \begin{equation}
% \label{eq: EI_s}
% \text{EI}_s(\vx) = \mathbb{E}[\max(\nablafnorm_2^2 - \|\nabla f(\vx^+)\|_2^2, 0)]. 
% \end{equation}

The \textcolor{black}{lower bound} \eqref{eq: EI-GN lower bound} retains the standard EI mechanism via EI$_f$, but is conservative with respect to stationarity. This is due to EI$_s$ only penalizing upward excursions of $\|\nabla f(\vx)\|_2^2$ relative to the incumbent and not directly rewarding gradient norm decreases. In low-improvement regions where $\mathrm{EI}_f(\vx)\approx 0$, maximizing the $\text{EI}_g$ \textcolor{black}{lower bound} \eqref{eq: EI-GN lower bound} is approximately equivalent to minimizing $\mathrm{EI}_s(\vx)$, which implicitly biases search toward
near-stationary regions where improvement via stationarity is plausible (pathway (ii)).
% Since computing $\text{EI}_g(\vx)$ is intractable, we use the lower bound $\mathcal{A}(\vx)$ in practice. For convenience, we restate the EI-GN lower bound $\mathcal{A}(\vx) = \text{EI}_f(\vx) -\alpha \text{EI}_s(\vx)$. Therefore, this retains the standard EI mechanism via $\text{EI}_f$. But, it is conservative with respect to stationarity: EI-GN lower bound penalizes increases in $\nablafnorm_2^2$ but does not directly reward decreases. We note in regions where $\mathrm{EI}_f(\vx) \approx 0$, maximizing $\mathcal{A}$ is approximately equivalent to minimizing $\text{EI}_s(\vx)$, hence this biases search towards near-stationary regions where pathway (ii) is plausible.
\color{black}
\subsection{Regimes of Effectiveness}
\label{sec: regimes_of_effectiveness}
We provide intuition on when EI$_g$ is expected to be most effective. In the idealized case of an interior maximizer, Polyak--\L{}ojasiewicz can be invoked to show $\|\nabla f(\vx)\|_2^2 \geq 2\beta (f(\vx^\star) - f(\vx))$, where $\beta > 0$. %In other words, the gradient norm provides a proxy for suboptimality.
Hence, reducing $\|\nabla f(\vx)\|_2^2$ can correspond to progress toward the global optimum, and EI$_g$'s stationarity-aware formulation can be beneficial in smooth landscapes, particularly in flat or near-stationary regions where standard EI provides little signal. However, when local optima are themselves near-stationary (i.e., $\|\nabla f(\vx)\|_2 \approx 0$ even for suboptimal $\vx$), the gradient norm may no longer distinguish between basins.

% If the global maximizer $\vx^\star$ lies in the interior of $\mathcal{X}$, then $\nabla f(\vx^\star) = 0$ and hence $g(\vx^\star) = f(\vx^\star)$, suggesting that optimizing $g$ can be aligned with optimizing $f$.

% \begin{equation}
%     \label{eq:pl_transfer}
%     r_T(f) \lesssim \frac{1}{1 + 2\alpha\mu} \, r_T(g),
% \end{equation}
% where $r$ denotes regret. This suggests that $f$-regret may be smaller than $g$-regret by a factor depending on the penalty weight $\alpha$ and the PL constant $\mu$. For $\alpha = 0$, $g=f$ and the regrets coincide.

% This interpretation highlights that the gradient norm provides a proxy for suboptimality.

% This discussion is intended to provide intuition rather than a formal theoretical guarantee. A more rigorous analysis of regret and a precise characterization of regimes where EI$_g$ is advantageous are left for future work.
\color{black}

\subsection{Expected Improvement via Gradient Norms}
\label{sec: EIGN lower bound approx}
Although the \textcolor{black}{lower bound} \eqref{eq: EI-GN lower bound} reduces the intractability of $\mathrm{EI}_g$ to evaluating the gradient term $\mathrm{EI}_s$ in \eqref{eq: EI_s}, $\mathrm{EI}_s$ remains impractical to compute within acquisition optimization. We therefore introduce a tractable \textcolor{black}{approximation} $\overline{\mathrm{EI}}_s$ and use it to define the practical acquisition EI-GN. 

To this end, we first define $\cholesky$ as the Cholesky decomposition for a positive definite covariance $\Sigmanabla \succ 0$ such that $\Sigmanabla = \cholesky\cholesky^\top$. Hence, we rewrite $\nablaf$ via the whitening transformation: $\nablaf = \munabla + \cholesky \vz$, where $\vz \sim \mathcal{N}(\mathbf{0}, \mathbf{I})$. This also permits a change of variables $\vz^+ = \cholesky^{-1}(\nabla f(\vx^+) - \munabla)$. Recall the independent modeling assumption from Section~\ref{sec: model}, which yields diagonal $\Sigma^\nabla(\vx)$ and $\mathbf{L}$, with distinct elements. This leads to the case where $\nablafnorm_2^2$ becomes a generalized noncentral chi-square variable
\begin{equation}
\label{eq: EI_s chi square}
    \text{EI}_s(\vx) = \int_{\|\nabla f(\vx^+)\|^2_2}^{\infty}\left(u - \|\nabla f(\vx^+)\|_2^2\right)p_{g\chi'^2}(u)du,
\end{equation}
where $u$ denotes the random variable $\nablafnorm_2^2$. Computing \eqref{eq: EI_s chi square} is nontrivial since it does not have a closed-form. Methods such as characteristic function inversion \cite{imhof1961computing, davies1980algorithm} use iterative numerical integration and are impractical to invoke repeatedly within AF optimization.

We recognize the difficulty of evaluating \eqref{eq: EI_s chi square} stems from the non-separable truncation of $u = \|\munabla + \cholesky\vz\|_2^2$. \textcolor{black}{Therefore, we address this by decoupling the norm-based truncation into a separable form that enables tractable evaluation of the expectation on a per-dimension basis.} We now define $\overline{\text{EI}}_s$ as the \textcolor{black}{tractable form} of $\text{EI}_s$
\begin{equation}
\label{eq: EI_s overline}
    \overline{\text{EI}}_s(\vx) = \int_{\vz^+}^{\infty}
		\left(\|\munabla + \cholesky\vz\|_2^2 - \|\nabla f(\vx^+)\|_2^2\right)\phi_{\vz}(\vz)d\vz,
\end{equation}
where $\phi_{\vz}$ denotes the multi-variate PDF of $\vz$. This yields separable truncated moments and a closed-form expression; see Appendix~\ref{app: EI-GN lower bound approx} for the full derivation. \textcolor{black}{In practice, we optimize this tractable form as a proxy for the lower bound in \eqref{eq: EI-GN lower bound}, yielding the final acquisition EI-GN}
\begin{equation}
\label{eq: EI-GN final}
    \text{EI-GN}(\vx) = \text{EI}_f(\vx) -\alpha\overline{\text{EI}}_s(\vx).
\end{equation}
%$\text{EI-GN}(\vx) = \text{EI}_f(\vx) -\alpha\overline{\text{EI}}_s(\vx)$. 
The computational complexity of \eqref{eq: EI-GN final} is discussed in Appendix~\ref{app: complexity}, and its pseudocode is given in Algorithm~\ref{alg: EI-GN}. % Although Section~\ref{sec:acquisition_signal} analyzes the $\text{EI}_g$ lower bound, EI-GN is derived as a tractable approximation and is designed to preserve the same stationarity bias characterized there.
Beyond tractability, the construction maintains the stationarity bias by penalizing large gradients, but this is achieved on a per-dimension basis. 
%In other words, we alter the order of operations: the positive-part truncation is first applied on per-dimension contributions then aggregated. This contrasts the norm-based generalized $\chi^2$ approach, which aggregates across dimensions before applying truncation on the scalar $u$. The reordering induces a \textcolor{blue}{per-dimension stationarity bias}; the quadratic norm aggregation may allow one coordinate's contribution to $u$ dominate, whereas the per-dimension approach places more emphasis on distributing the stationarity penalty across all coordinates. 
\textcolor{black}{We provide empirical evidence that this yields similar performance in Appendix~\ref{sec: orthant_approximation}.} %This approximation may deviate from the norm-based formulation in highly anisotropic settings, where a small subset of dimensions dominates the gradient norm, potentially leading to discrepancies in candidate ranking.} 

\section{Experiments}

\subsection{Experimental Setup}
\textcolor{black}{
We perform different experiments with varying baselines: % to answer the following questions:
\begin{enumerate}
    \item Synthetic Benchmarks (Section~\ref{sec: synthetic}): EI-GN is compared with a comprehensive set of baselines to establish regimes of effectiveness (Section~\ref{sec: regimes_of_effectiveness}) empirically.
    \item GP Samples (Section~\ref{sec: gp_samples}): Acquisition behavior is isolated with comparisons against functions in the EI-family. This ascertains the usefulness of stationarity-aware objectives \eqref{eq: g} for improvement based AFs and the alternative pathway to improvement via gradient norms (Section~\ref{sec:acquisition_signal}).
    \item Gradient Ablation (Section~\ref{sec: gradient_ablation}): Gradient-based baselines are compared against EI-GN. Specifically, we show surrogates that incorporate gradients are orthogonal to the proposed method and can yield  complementary gains when combined with EI-GN.
    \item Application to Policy Search (Section~\ref{sec: policy_search}): EI-GN's performance is demonstrated in policy search case studies, where gradients are estimated via the policy gradient theorem. We contrast local gradient-based methods (i.e. REINFORCE) with global methods (i.e. EI-GN) when gradients are available.
\end{enumerate}
}

Each experiment is repeated across 20 seeds, initialized with $3d$ quasi-random Sobol points. All results are shown with mean $\pm$ one standard error of best $f$. Unless stated otherwise, we model GPs with the Mat\'ern-$5/2$ kernel along with Automatic Relevance Determination (ARD); additional implementation details can be found in Appendix~\ref{app: implementation}. Here, the function and gradient observations are assumed to be deterministic to isolate acquisition behavior without confounding due to objective noise. We find that EI-GN performance is largely insensitive to $\alpha$ and use $\alpha = 0.6$ for all EI-GN experiments; hyperparameter tuning experiments for $\alpha$ can be found in Appendix~\ref{app: hyperparameter alpha}. \textcolor{black}{The code implementation of EI-GN is available at:
\url{https://github.com/ipjoshua1483/EI-GN}.}

% Additionally, we perform controlled ablations to empirically verify when EI-GN outperforms; within/out-of-model experiments show EI-GN against EI-variants when surrogate fitting is removed from BO and gradient-based comparisons ascertain whether gradients at the surrogate level are beneficial. We also explore extentions of EI-GN to policy search, where noisy objective and gradient observations are assumed. 

% The supplementary material includes (i) a minimal code implementation and (ii) the data and plotting scripts to generate Figures 3, 8, and 10.

% We use EI as the primary EI-family baseline since it is canonical and directly connected to our formulation (e.g., the $\alpha=0$ special case) and defer comparisons to LogEI \cite{ament2023unexpected}, a variant designed to mitigate EI's vanishing acquisition signal, to Appendix~\ref{app: EI variants}. 

% Comparisons of EI-GN to derivative-free baselines are not information-matched since we assume $(y,\nabla y)$ is observed per query, and our goal is to study how to exploit gradients when they are available at no extra evaluation cost.

% In addition to benchmarking, we also perform diagnostics on policy learning under global and local optimization perspectives; we compare EI-GN with REINFORCE \cite{williams1992simple} as representatives of each framework respectively. Since the inherent nature of environment dynamics naturally leads to stochastic systems, we fix seeds and use averaged returns across a sufficiently large number of rollouts to surpress noise.

\subsection{Synthetic Benchmarks}
\label{sec: synthetic}
\begin{figure*}[tb]
    \centering
    \includegraphics[width=\textwidth]{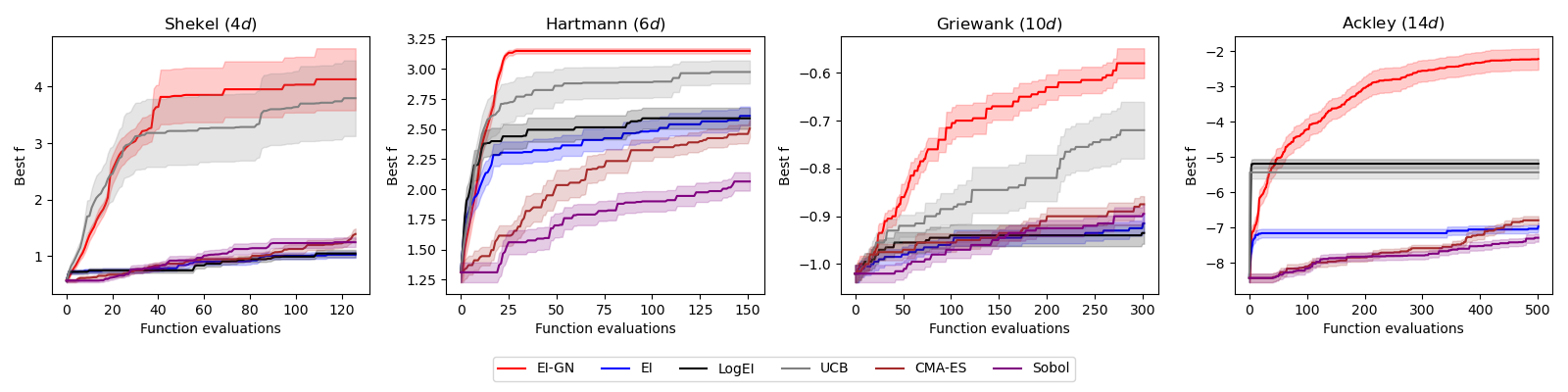}
    \caption{Results displaying mean $\pm$ one standard error for synthetic benchmarks: Shekel (4$d$), Hartmann (6$d$), Griewank (10$d$), and Ackley (14$d$) for EI-GN against a variety of baselines.}
    \label{fig: synthetic}
\end{figure*}
\textcolor{black}{We first evaluate overall performance using a comprehensive set of BO baselines.} We analyze performance for synthetic benchmarks across a range of dimensions: Shekel (4$d$), Hartmann (6$d$), Griewank (10$d$), Ackley (14$d$). These problems are standard in optimization literature and are commonly used to establish performance \cite{surjanovic2014virtual}. We benchmark the proposed method against several synthetic functions; we compare EI-GN against EI, \textcolor{black}{LogEI, Upper Confidence Bound (UCB) \cite{srinivas2009gaussian}}, CMA-ES \cite{hansen2001completely}, and Sobol.

Figure~\ref{fig: synthetic} displays the results of EI-GN against the other baselines for the synthetic benchmarks. For Shekel (4$d$), EI-GN outperforms all other baselines except for UCB, where the results are similar. This can be explained by the function exhibiting multiple well-defined basins where exploration-based methods such as UCB can effectively identify promising regions, leading to performance comparable to EI-GN. \textcolor{black}{For Hartmann (6$d$), both EI-GN and EI have near identical performance in the first few BO iterations, but the latter stagnates soon after due to the improvement criterion providing insufficient signal. EI-GN provides additional gains in this setting, indicating that the standard improvement signal is no longer sufficient. In Griewank (10$d$) and Ackley (14$d$), EI-GN significantly outperforms the other baselines, since the functions exhibit oscillatory or flat structure that offer weak improvement signals, thereby highlighting the importance of gradient information in these contexts.}  %Furthermore, the independent gradient-enhanced GPs ignore correlations and only yield approximate gradient inference, meaning the stationarity term in EI-GN is not operating under fully accurate gradient information, yet this still effectively guides optimization in search space. \textcolor{blue}{Notably, UCB significantly underperforms compared to EI-GN.} EI appears to have similar initial performance in Ackley (14$d$), but soon collapses to a local maxima. 

% In Griewank (10$d$) and Ackley (14$d$), EI-GN significantly outperforms all baselines. These functions exhibit oscillatory or flat structure with weaker improvement signals, where exploration alone is insufficient, and EI-GN provides more consistent gains.

\textcolor{black}{Across all four synthetic benchmarks, LogEI underperforms compared to EI-GN and mostly has similar performance to EI. This demonstrates the differences in how these two EI-based acquisitions operate; LogEI modifies EI term to prevent collapse due to vanishing values, whereas EI-GN injects objectives with stationarity, which provides an additional pathway to improvement via the gradient term \eqref{eq:g_decomposition}. This behavior is consistent with the intuition in Section~\ref{sec: regimes_of_effectiveness}: EI-GN provides the largest gains in regimes where the standard improvement signal becomes weak, such as flat or near-stationary regions, while offering limited advantage when improvement signals are already strong.}

% The relative performance of each method is heavily affected by multimodality and flat regions of the objective function, and is not merely a function of dimensionality. As shown in synthetic benchmarks of moderate dimension (i.e., Hartmann (6$d$), Cosine (8$d$)), exploration-based AFs such as TS are competitive to avoid over-exploitation. However, they are insufficient for more complex objective landscapes (i.e. Griewank (10$d$), Ackley (14$d$)) where gradient information is necessary to effectively guide optimization in search space. Furthermore, the independent gradient-enhanced GPs ignore correlations and only yield approximate gradient inference, meaning the stationarity penalties are not operating under fully accurate gradient information. % Ablations and hyperparameter analysis for these synthetic benchmarks can be found in Appendix~\ref{app: ablation_synthetic}.
\subsection{GP samples}
\label{sec: gp_samples}
\textcolor{black}{We restrict to EI-family baselines to isolate acquisition-level effects.} In GP samples, we consider both within-model  comparisons \cite{hennig2012entropy} and out-of-model comparisons \cite{muller2021local} in 7$d$, 8$d$, and 9$d$. \textcolor{black}{Here, we compare EI-GN with other acquisition functions in the EI family: EI, LogEI, and EI with exploration (EI$_\xi$). We evaluate EI$_\xi$ in $\xi \in [0.01, 0.1, 1.0]$. This controlled comparison isolates the effect of modifying the improvement criterion within a common acquisition framework.} 

\subsubsection{Within-model Comparisons}
We analyze performance under within-model comparisons \cite{hennig2012entropy} to establish acquisition performance under idealized assumptions where model structure and hyperparameter mismatches are eliminated. This is achieved with objectives sampled via GP priors and can be considered the best-case regime for AFs of interest. This isolates the impact of gradient information in EI-GN to ascertain whether the stationarity penalty effectively guides optimization even when models are specified perfectly. In addition, the GP for $f$ is blind to the gradient observations due to the structure of the gradient-enhanced GPs, therefore any performance gains must be solely due to the gradient-based term. We sample objectives from GP priors with RBF kernels and use the same kernel class in the surrogates to preserve matched model specification.

\begin{figure}[b!]
    \centering
    \includegraphics[width=\columnwidth]{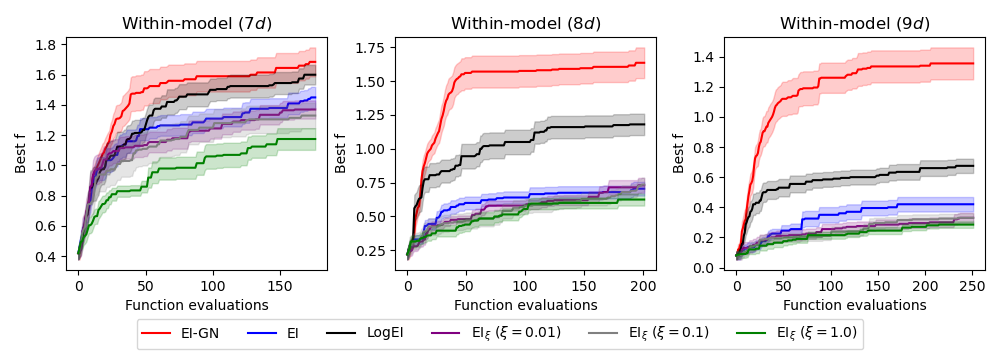}
    \caption{Results displaying mean $\pm$ one standard error for within-model comparisons in 7$d$, 8$d$, and 9$d$.}
    \label{fig: within model}
\end{figure}

Figure~\ref{fig: within model} shows EI-GN outperforming the other EI-based baselines, suggesting acquisition behavior can be directly improved with gradient information to inform stationarity rather than indirectly via improved surrogates. The other baselines achieve similar results to EI-GN for 7$d$, but their performances degrade with 8$d$ and 9$d$. This reinforces the prior claim that \textcolor{black}{EI-GN has superior performance due to the additional pathway of improvement. In many problem landscapes, the improvement criterion will not provide adequate signal once a sufficiently large incumbent is found, causing EI to fail. Furthermore, scaling to the EI term or additional exploration with EI$_\xi$ are also insufficient in this context. However, improvement derived from stationarity-aware objectives gives additional signal and provides beneficial modifications to the acquisition lanscape.}

%EI over-exploits, as even under perfect surrogate assumptions, sufficiently complex problem landscapes will eventually cause premature convergence to local maxima. Synthetic benchmarks tend to have many high-valued local maxima, which favor exploration via posterior sampling. However, objectives sampled from GP priors may be multimodal, yet only contain relatively few high-reward basins; hence, TS suffers from these objective landscapes. In these regimes, structured exploration via stationarity penalties is more impactful than exploration via posterior sampling. % Ablations and hyperparameter analysis for within-model comparisons can be found in Appendix~\ref{app: within}.

\subsubsection{Out-of-model Comparisons}
% This considers model mismatch in addition to within-model comparisons \cite{hennig2012entropy}, where both the objective and model are from identical settings. Out-of-model comparisons differ from traditional synthetic benchmarks since they maintain smooth GP-like objective landscapes while deviating from ideal model assumptions, allowing analysis of acquisition performance under model mismatch and hyperparameter optimization. This contrasts synthetic benchmarks which often contain multiple sources of mismatch, allowing isolated analysis of acquisition behavior.

Unlike within-model comparisons, the hyperparameters of the GP prior used to generate the objective and those in the surrogate for BO vary for out-of-model comparisons \cite{muller2021local}. This allows investigation of acquisition performance under model mismatch and hyperparameter optimization. However, this differs from traditional synthetic benchmarks since they maintain smooth GP-like objective landscapes while deviating from ideal model assumptions. Therefore, out-of-model comparisons provide a controlled, yet more realistic environment for benchmarking EI-GN, where the effect of stationarity penalties is investigated under smooth objectives (and gradients) with imperfect surrogates. We hold all experimental conditions from the within-model comparisons constant, except we revert back to the Mat\'ern-$5/2$ kernel with ARD to introduce kernel mismatch.

\begin{figure}[h]
    \centering
    \includegraphics[width=\columnwidth]{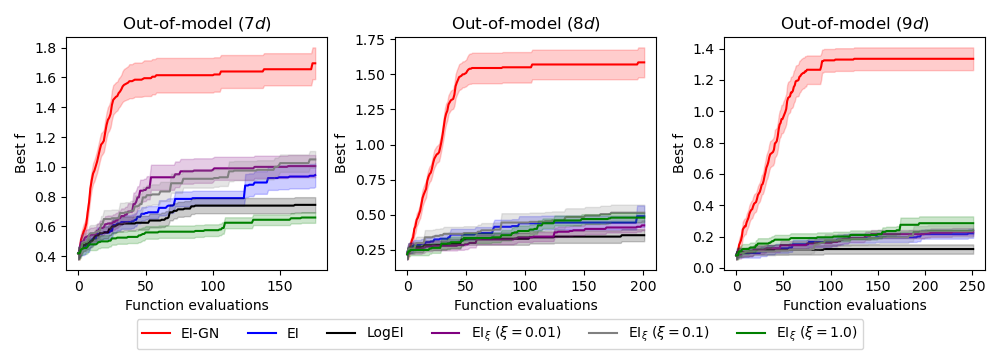}
    \caption{Results displaying mean $\pm$ one standard error for out-of-model comparisons in 7$d$, 8$d$, and 9$d$.}
    \label{fig: out of model}
\end{figure}

In Figure~\ref{fig: out of model}, not only do we observe strong EI-GN results, its performance compared to the within-model comparisons is largely unchanged, highlighting the importance of gradient information. Furthermore, this also demonstrates that they are robust to approximate posteriors derived from uncorrelated surrogates and model misspecifications. \textcolor{black}{Accordingly, EI underperforms, and additional exploration via EI$_\xi$ yields similar conclusions. LogEI performs similarly since the root cause for EI's subpar performance is due to model misspecification, meaning numerical scaling will not provide any improvements.}

% On the other hand, EI underperforms and is on par with TS. When only informed on zeroth-order information, inaccurate beliefs of problem landscape can cause convergence to even more suboptimal basins. Although TS underperforms as well, its performance largely remains constant from the within model case. This can be attributed to it being less sensitive to model mismatches due to posterior sampling exploration; yet it retains similar performance to the within-model comparisons due to the same failure modes. % Ablations and hyperparameter analysis for out-of-model comparisons can be found in Appendix~\ref{app: out of model}.

In both within and out-of-model comparisons, EI-GN consistently outperforms all other baselines, which can be attributed to the smooth, GP-like objectives leading to higher quality gradient observations. The stationarity penalty in EI-GN effectively drives optimization towards relevant regions of the search space, thereby speeding up discovery of desired maxima, even when uncorrelated models are approximate and offer no improvements to surrogate accuracy in $f$, as well as model mispecifications. Other acquisition functions must rely on zeroth order information only and cannot take advantage of gradient structure, triggering performance degradations with landscape complexity and model mismatches. 
\color{black}
\subsection{Gradient Ablation}
\label{sec: gradient_ablation}
\textcolor{black}{We compare gradient-based methods to isolate the role of gradient information.} In this work, we focus on incorporating gradients at the acquisition level, which results in the proposed EI-GN. As motivated in Sec.~\ref{sec: model}, the choice of independent models allows us to isolate improvements in performance to the acquisition, since the surrogate itself does not directly benefit from gradient observations. This also reduces the overall complexity from $\mathcal{O}((d+1)^3N^3)$ to $\mathcal{O}((d+1)N^3)$.

\begin{table}[h]
\centering
\caption{Comparison of how gradient information is incorporated across methods.}
\begin{tabular}{|l|c|c|c|c|}
\hline
 Presence of gradients & dEI-GN & dEI & EI-GN & EI \\
\hline
Joint GP        & \cmark & \cmark & \xmark & \xmark \\
\hline
Independent GP & \xmark & \xmark & \cmark & \xmark \\
\hline
Acquisition function    & \cmark & \xmark & \cmark & \xmark \\
\hline
\end{tabular}
\label{tab:gradient_usage}
\end{table}

Gradient-enhanced GP methods (e.g., joint models with derivative observations \cite{wu2017bayesian}) instead incorporate gradient information at the surrogate level by improving posterior estimation of $f$, which is orthogonal to EI-GN. As a result, these two approaches can be combined to yield a complementary method, dEI-GN. Table~\ref{tab:gradient_usage} summarizes the presence of gradients in dEI-GN, dEI, EI-GN, and EI. We restrict this analysis to lower-dimensional settings due to the cubic scaling of gradient-enhanced Gaussian process models with derivative observations, which makes joint modeling prohibitively expensive in higher dimensions.

\begin{figure}[h]
    \centering
    \includegraphics[width=\columnwidth]{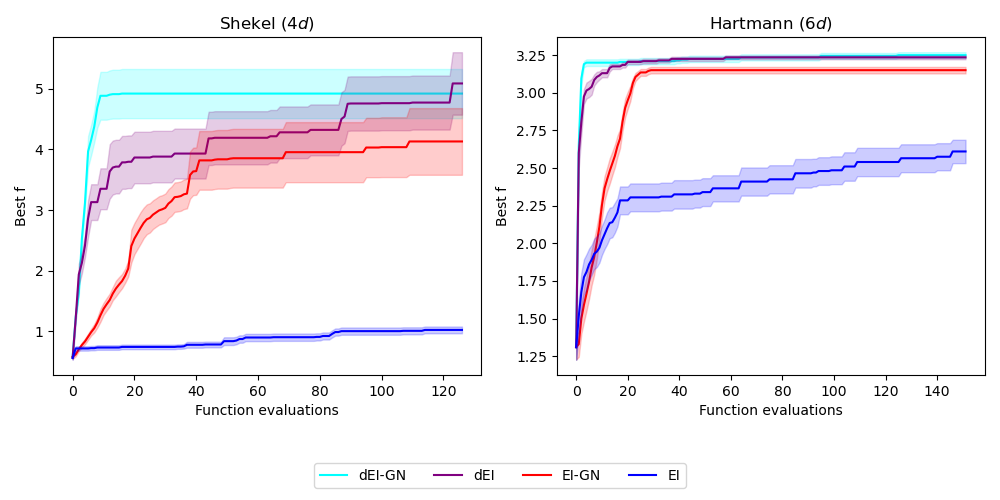}
    \caption{Mean $\pm$ one standard error for Shekel (4$d$) and Hartmann (6$d$) displayed for dEI-GN, dEI, EI-GN, EI.}
    \label{fig: gradient_baseline}
\end{figure}

Figure~\ref{fig: gradient_baseline} isolates the effect of incorporating gradient information at different stages. Gradient-enhanced methods such as dEI improve performance by refining the surrogate model through derivative observations, while EI-GN improves performance by modifying the acquisition to incorporate stationarity-aware signals. \textcolor{black}{These correspond to two orthogonal axes of improvement: posterior modeling and acquisition design.} In Shekel (4$d$), both axes contribute to performance gains, and their combination (dEI-GN) yields the strongest results, demonstrating complementarity. In Hartmann (6$d$), dEI rapidly achieves near-optimal performance, leaving limited room for further gains from acquisition design. In this regime, dEI-GN performs comparably to dEI, indicating that improvements from acquisition design are most impactful when the surrogate alone is insufficient. These results highlight that acquisition-level and surrogate-level improvements address complementary aspects of the optimization problem.

\color{black}
\subsection{EI-GN Application to Policy Search}
\label{sec: policy_search}
\begin{figure}[tb]
    \centering
    \includegraphics[width=\columnwidth]{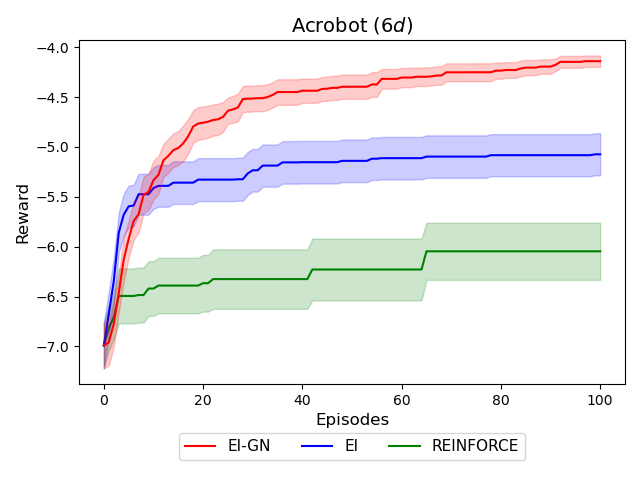}
    \caption{Mean $\pm$ one standard error for Acrobot (6$d$) displayed for EI-GN (red), EI (blue), and REINFORCE (green).}
    \label{fig: acrobot}
\end{figure}
\begin{figure}[tb]
    \centering
    \includegraphics[width=\columnwidth]{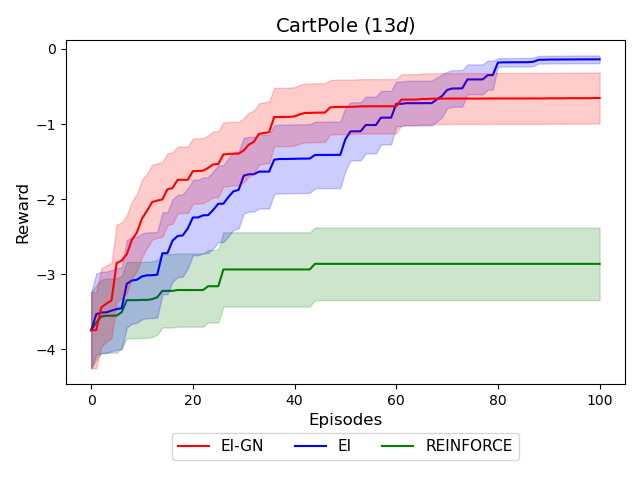}
    \caption{Mean $\pm$ one standard error for Cartpole (13$d$) displayed for EI-GN (red), EI (blue), and REINFORCE (green).}
    \label{fig: cartpole}
\end{figure}

% We seek to evaluate EI-GN on more realistic problems where prior assumptions on objective and gradient observations are relaxed; policy search problems in the reinforcement learning literature are the ideal testbeds for our proposed method. The objectives are typically noisy due to stochasticity in the environment dynamics, and an average reward is usually computed across rollouts. Since gradients of the expected reward cannot be directly observed, they must be estimated via approaches such as policy gradient theorem \cite{sutton1999policy}.

\textcolor{black}{This experiment compares global BO methods and local policy gradient methods when gradients are available.}
%We include EI as a representative gradient-free global baseline, EI-GN as the proposed gradient-aware global method, and REINFORCE as a representative local gradient-based approach.} 
We now demonstrate EI-GN on solving policy search problems. Policy search methods aim to learn optimal parameters of a stochastic policy using noisy observations of a reward function. Since gradients of the expected reward are not directly observable, they are estimated using the Policy Gradient Theorem \cite{sutton1999policy}.

Policy search is often performed using local stochastic gradient methods that rely on gradient directions and lack mechanisms for global exploration. BO has been used in policy search with strong sample efficiency, yet it treats the expected reward as a black-box objective. In contrast, trajectory-level signals can be exploited to improve policy search \cite{wilson2014using}. We position EI-GN as a hybrid approach that combines BO’s global exploration with gradient-based policy learning by directly incorporating policy gradient estimates—computed from the same rollouts used to estimate rewards—when proposing new policies. We compare EI-GN to EI and REINFORCE \cite{williams1992simple} as representative zeroth- and first-order methods, respectively, and evaluate performance on a linear policy in Acrobot (6$d$) \cite{sutton1995generalization} and a neural policy in Cartpole (13$d$) \cite{barto1988neuronlike}.

% Policy learning is typically performed with stochastic gradient methods (i.e. actor-critic, PPO). These methods are inherently local and primarily use gradient information as parameter update directions; they do not explicitly reuse historical evaluations to guide global exploration for optimal policies. On the other hand, BO has been explored in policy search and can be sample efficient for expensive policy evaluations, but most works treat expected return as the black-box objective. In contrast, trajectory-derived signals can be leveraged to improve policy search as shown by \citet{wilson2014using}. 

% Therefore, we view EI-GN as a hybrid optimization method that combines the global perspective of BO with gradient-based policy learning approaches by directly using policy gradient estimates (past and present) when suggesting new policies. Importantly, the gradient estimates can be computed from the same rollouts used to estimate expected reward without additional environment interaction via the policy gradient theorem. We compare EI-GN against EI and REINFORCE \cite{williams1992simple} as representative global (zeroth-order) and local (first-order) methods, respectively. We show results for a linear policy in Acrobot (6$d$) \cite{sutton1995generalization} and a neural policy in Cartpole (13$d$) \cite{barto1988neuronlike}. % To alleviate variance of gradient estimates derived from the policy gradient theorem, we subtract a baseline comprised of the average reward over all rollouts.

As seen in Figure~\ref{fig: acrobot}, EI-GN and EI have similar performance on Acrobot in early iterations, but EI then plateaus, suggesting sensitivity to local basins. EI-GN exhibits lower variance across seeds and reliably reaches higher rewards. For Cartpole shown in Figure~\ref{fig: cartpole}, EI-GN and EI achieve similar performance, which is attributed to the reduced informativeness of policy gradients due to increased policy complexity and consequently higher-variance gradient estimation. In both problems, EI-GN is more sample-efficient than REINFORCE, but this comparison is intended only as a contrast to purely local first-order updates. % rather than superiority in reinforcement learning performance. 
We recognize that policy search methods are sensitive to design choices (i.e., learning rates, entropy regularization, gradient clipping).

\section{Conclusion}

We build upon EI by applying the improvement criterion to a gradient-aware auxiliary variable, \textcolor{black}{which provides an additional pathway to improvement beyond objective value}. We derive a new acquisition function and present a closed-form approximation, EI-GN, for its computation. Through empirical evaluation on synthetic benchmarks, GP sample-based objectives, \textcolor{black}{gradient ablations}, and policy search problems, we demonstrate that EI-GN achieves competitive performance. Future work will extend EI-GN to improve robustness under noisy objectives and gradient estimates \textcolor{black}{via variance-reduction techniques (i.e., value/advantaged-baselines, generalized advantage estimation, natural policy gradients) without increasing sample complexity.}

\bibliography{example_paper}
\bibliographystyle{icml2026}

%%%%%%%%%%%%%%%%%%%%%%%%%%%%%%%%%%%%%%%%%%%%%%%%%%%%%%%%%%%%%%%%%%%%%%%%%%%%%%%
%%%%%%%%%%%%%%%%%%%%%%%%%%%%%%%%%%%%%%%%%%%%%%%%%%%%%%%%%%%%%%%%%%%%%%%%%%%%%%%
% APPENDIX
%%%%%%%%%%%%%%%%%%%%%%%%%%%%%%%%%%%%%%%%%%%%%%%%%%%%%%%%%%%%%%%%%%%%%%%%%%%%%%%
%%%%%%%%%%%%%%%%%%%%%%%%%%%%%%%%%%%%%%%%%%%%%%%%%%%%%%%%%%%%%%%%%%%%%%%%%%%%%%%
\newpage
\appendix
\onecolumn
\section{Positive Part Inequality}
\label{app: positive part inequality}
\begin{lemma}
\label{lemma: positive part inequality}
For real-valued $A$ and $B$,
    \begin{equation}
        \max(A-B, 0) \geq \max(A, 0) - \max(B, 0).
    \end{equation}

\begin{proof}
We perform case analysis on the signs of $A, B$.

\textbf{Case 1:} $A \geq 0$, $B \geq 0$, $\max(A,0) - \max(B,0) = A - B$.\\
If $A - B \geq 0$, $\max(A - B,0) = A-B$, hence inequality holds with equality.\\
If $A - B < 0$, $\max(A - B,0)=0 > A-B$, hence inequality holds.

\textbf{Case 2:} $A \geq 0$, $B < 0$, $\max(A,0) - \max(B,0) = A$.\\
$\max(A-B,0) = A-B$. Since $B < 0$, $A - B > A$, hence inequality holds.

\textbf{Case 3:} $A < 0$, $B \geq 0$, $\max(A,0) - \max(B,0) = -B \leq 0$. \\
$\max(A-B,0)=0$, hence inequality holds.

\textbf{Case 4:} $A < 0$, $B < 0$, $\max(A,0) - \max(B,0) = 0$. \\
If $A < B$, then $A - B < 0$ and $\max(A - B, 0) = 0$, hence the inequality holds with equality. \\
If $A \geq B$, then $A - B \geq 0$ and $\max(A - B, 0) = A - B \geq 0$, hence the inequality holds.

\end{proof}

\end{lemma}
\section{$\text{EI}_g$ \textcolor{blue}{Lower Bound}}
\subsection{Derivation of $\text{EI}_g$ Lower Bound}
\label{app: EI-GN}
We proceed to derive the expected improvement (EI) acquisition function on $g(\vx)$.
\begin{subequations}
\begin{align}
\label{eq: EI_g_1}
    \text{EI}_g(\vx) &= \mathbb{E}\left[\max(g(\vx) - g(\vx^+), 0)\right]\\
    \text{EI}_{g}(\vx) &= \mathbb{E}\left[\max\big(\f - f(\vx^+) - \alpha\big(\nablafnorm_2^2 - \|\nabla f(\vx^+)\|_2^2\big), 0\big)\right]\\
    \text{EI}_{g}(\vx) &\geq \mathbb{E}\left[\max\big(\f - f(\vx^+), 0)\right] - \alpha\mathbb{E}\left[\max\big(\nablafnorm_2^2 - \|\nabla f(\vx^+)\|_2^2, 0\big)\right]
\end{align}
\end{subequations}

where \textcolor{blue}{an inequality} is applied when it is expressed as a sum of two max operators, based on Positive Part Inequality (Lemma~\ref{lemma: positive part inequality}). The first expectation corresponds to EI$_f$ 
\begin{equation}
\label{eq: EI_g_lower_bound}
    \text{EI}_g(\vx) \geq \text{EI}_f(\vx) - \alpha\mathbb{E}\left[\max\big(\nablafnorm_2^2 - \|\nabla f(\vx^+)\|_2^2, 0\big)\right]
\end{equation}
\section{EI-GN}
\subsection{Derivation of EI-GN}
\label{app: EI-GN lower bound approx}
We define $\cholesky$ to be the Cholesky decomposition for a positive definite covariance $\Sigmanabla \succ 0$ such that $\Sigmanabla = \cholesky\cholesky^\top$. Hence, we rewrite $\nablaf$ via the whitening transformation
\begin{equation}
    \label{eq: whitening}
    \nablaf = \munabla + \cholesky \vz
\end{equation}
where $\vz \sim \mathcal{N}(\mathbf{0}, \mathbf{I}_d)$. Therefore,

\begin{equation}
    \mathbb{E}\left[\max\big(\nablafnorm_2^2 - \|\nabla f(\vx^+)\|_2^2, 0\big)\right] =\mathbb{E}\left[
    \max\left(
    \|\munabla + \cholesky\vz\|^2_2 - \|\nabla f(\vx^+)\|_2^2, 0\right)\right]
\end{equation}
$\nablafnorm_2^2$ follows the generalized chi-square distribution
\begin{equation}
% \label{eq: EI_s chi square}
    = \int_{\|\nabla f(\vx^+)\|^2_2}^{\infty}\left(u - \|\nabla f(\vx^+)\|_2^2\right)p_{g\chi'^2}(u)du,
\end{equation}
where $u$ denotes the random variable $\nablafnorm_2^2$. We note that evaluation of this integral is prohibitively expensive, hence we approximation the norm-based truncation with the orthant truncation anchored at the incumbent (Section~\ref{sec: EIGN lower bound approx})
\begin{equation}
\Big\{ \|\mu^\nabla(\vx)+\cholesky\vz\|_2^2 \ge \|\nabla f(\vx^+)\|_2^2 \Big\}
\;\approx\;
\Big\{ \vz \geq_\text{cw} \vz^+ \Big\}
\end{equation}
% We define the multi-variate PDF of $\vz$ as $\phi_{\vz}$ and expand the expectation 
% \begin{equation}
%     =\int_{-\infty}^{\infty}
%     \max\left(\|\munabla + \cholesky\vz\|^2_2 - \|\nabla f(\vx^+)\|_2^2, 0\right)\phi_{\vz}(\vz)d\vz
% \end{equation}
% We now make a change of variables to refer to the $\vz^+$ that corresponds to $\nabla f(\vx^+)$. 
With \eqref{eq: whitening}, $\vz^+ = \cholesky^{-1}(\nabla f(\vx^+) - \munabla)$ and this yields

\begin{equation}
    =\int_{\vz^+}^{\infty}
\left(\|\munabla + \cholesky\vz\|^2_2 - \|\nabla f(\vx^+)\|_2^2\right)\phi_{\vz}(\vz)d\vz
\end{equation}

%It is recognized that the change of variables assumes that the gradient magnitude of a dimension must improve for a smaller norm, and this is an approximation to make further derivation tractable. 

This can be expanded into 3 terms

\begin{subequations}
\label{eq: three_terms}
    \begin{align}
    \label{eq: three_terms_1}
        &= \int_{\vz^+}^{\infty}(\munabla^\top \munabla - \|\nabla f(\vx^+)\|_2^2)\phi_{\vz}(\vz)d\vz\\
    \label{eq: three_terms_2}
        &+ 2\int_{\vz^+}^{\infty}\munabla^\top \cholesky \vz\phi_{\vz}(\vz)d\vz\\
    \label{eq: three_terms_3}
        &+\int_{\vz^+}^{\infty}\vz^\top\Sigmanabla \vz\phi_{\vz}(\vz)d\vz
    \end{align}
\end{subequations}

Since $\vz$ follows a standard normal multi-variate, the PDF $\phi_{\vz}(\vz)$ can be expressed as a product of individual single-variate PDFs $\phi$
\begin{equation}
    \phi_{\vz}(\vz) = \prod_{i=1}^d \phi(z_i)
\end{equation}
Subsequently, Fubini's theorem can be applied in integration accordingly, where
\begin{equation}
    \int \phi_\vz(\vz)d\vz = \int \prod_{i=1}^d \phi(z_i)dz_i = \prod_{i=1}^d \int \phi(z_i)dz_i
\end{equation}

The constants can be taken out of the integral for the evaluation of \eqref{eq: three_terms_1}
\begin{subequations}
    \begin{align}
    % &\mathbb{E}[(\nabla \mu (x) - \nabla f(x^*))^\top(\nabla \mu(x)-\nabla f(x^*))]\\
    &=(\munabla^\top \munabla - \|\nabla f(\vx^+)\|_2^2)\int_{\vz^+}^{\infty}\phi_{\vz}(\vz)d\vz\\
    &=(\munabla^\top \munabla - \|\nabla f(\vx^+)\|_2^2)\prod_{i=1}^d\int_{z_i^+}^{\infty}\phi(z_i)dz_i\\
    &=(\munabla^\top \munabla - \|\nabla f(\vx^+)\|_2^2)\prod_{i=1}^d\Phi(-z_i^+)
    %(1-\Phi_i(z_i^*))
    \end{align}
\end{subequations}
where $\Phi$ denotes the single-variate CDF of a Gaussian distribution. As for \eqref{eq: three_terms_2}, we begin by factoring the constants w.r.t integration
\begin{subequations}
    \begin{align}
        2\munabla^\top \cholesky\int_{\vz^+}^{\infty}\vz\phi_{\vz}(\vz)d\vz
    \end{align}
\end{subequations}
% Since $z$ is the standard multivariable normal variable, the PDF can be rewritten as a product of independent univariable normal variables
% We express the integral as a product of integrals via Fubini's theorem
% \begin{equation}
%     \int_{\vz^*}^{\infty}\vz\phi_{\vz}(\vz)d\vz = \prod_{i=1}^d\int_{z_i^*}^\infty \vz\phi(z_i)dz_i%= \prod_{i=1}^d\int_{z_i^*}^{\infty}z_i\phi_i(z_i)dz_i
% \end{equation}

We evaluate the integral component-wise. For any $j\in[d]$,
\begin{equation}
\label{eq:trunc_first_moment_component}
\Big[\int_{\vz^+}^{\infty}\vz\,\phi_{\vz}(\vz)\,d\vz\Big]_j
= \int_{\vz^+}^{\infty} z_j \,\phi_{\vz}(\vz)\, d\vz .
\end{equation}

We extract $\phi(z_j)$ and evaluate the corresponding integral separately, but the other terms correspond to the integral of the PDF
\begin{subequations}
    \begin{align}
    &= \int_{z^+_j}^{\infty}z_j \phi(z_j)dz_j \times \prod_{i \neq j} \int_{z_i^+}^\infty \phi(z_i)dz_i\\
    &=\phi(z_j^+)\prod_{i \neq j}\Phi(-z_i^+)
    \end{align}
\end{subequations}
We now define $\vphi(\vz^+) = \left[\phi(z_1^+),\ldots,\phi(z_d^+)\right]^\top$ and $\vPhi(-\vz^+) = \left[\Phi(-z_1^+), \dots, \Phi(-z_d^+))\right]^\top$ as the vectorized PDF and CDF for $\vz$. Subsequently, the element-wise division of these two vectors can be defined as
\begin{equation}
    \label{eq: w}
    \mathbf{w} = \vphi(\vz^+) \oslash \vPhi(-\vz^+)
\end{equation}
Therefore \eqref{eq: three_terms_2} can now be expressed as
\begin{equation}
    % 2\munabla^\top \cholesky \bigg(\vphi(\vz^*) \oslash \vPhi(\vz^*)\bigg) \prod_{i = 1}^d \Phi(-z_i^*)
    2\munabla^\top \cholesky \mathbf{w} \prod_{i = 1}^d \Phi(-z_i^+)
\end{equation}

\eqref{eq: three_terms_3} can be expressed as the summation across $i, j = 1, \dots, d$
\begin{equation}
\int_{\vz^+}^{\infty}\vz^\top\Sigmanabla \vz\phi_{\vz}(\vz)d\vz= \sum_{i,j=1}^d \Sigmanabla_{ij}\,\underbrace{\int_{\vz^+}^{\infty} z_i\,z_j\,\phi_{\vz}(\vz)\,d\vz}_{I_{ij}}
\end{equation}

The evaluation of the integral $I_{ij}$ falls under two categories: $i = j$, $ i \neq j$. For the first case where $i = j$,
\begin{equation}
\int_{\vz^+}^{\infty} z_i^2\,\phi_{\vz}(\vz)d\vz
= \prod_{i \neq k} \Phi(-z_k^+)
% \Bigl(\prod_{i\neq j} [1 - \Phi_j(z_j^*)]\Bigr)
\int_{z_i^+}^{\infty} z_i^2\,\phi(z_i)\,dz_i
\end{equation}
We proceed with integration by parts with
\begin{equation}
u = z_i, \quad dv = z_i\,\phi(z_i)\,dz_i 
\quad\Longrightarrow\quad 
du = dz_i, \quad v = -\phi(z_i)\notag
\end{equation}
\begin{subequations}
\begin{align}
\int_{z_i^+}^{\infty} z_i^2\,\phi(z_i)\,dz_i
&= \Bigl[-\,z_i\,\phi(z_i)\Bigr]_{z_i^+}^{\infty}
  + \int_{z_i^*}^{\infty} \phi(z_i)\,dz_i \\
&= z_i^+\,\phi(z_i^+) + \Phi(-z_i^+)
%\bigl[1 - \Phi_i(z_i^*)\bigr].
\end{align}
\end{subequations}
Therefore,
\begin{equation}
\int_{\vz^+}^{\infty} z_i^2\,\phi_{\vz}(\vz)d\vz = \left(z_i^+\,\phi(z_i^+) + \Phi(-z_i^+)\right)\prod_{i \neq k} \Phi(-z_k^+)
\end{equation}
The second case is where $i \neq j$
\begin{subequations}
\begin{align}
    \int_{\vz^+}^\infty z_iz_j\phi_{\vz}(\vz)d\vz 
    &= \int_{z_i^+}^\infty z_i\phi(z_i)dz_i \times \int_{z_j^+}^\infty z_j \phi(z_j)dz_j\times\prod_{k \neq i, j}\int_{z_k^+}^\infty \phi(z_k)dz_k\\
    &= \phi(z_i^+)\phi(z_j^+)\prod_{k \neq i, j}\Phi(-z_k^+)
    %(1-\Phi_k(z_k^*))
\end{align}
\end{subequations}

To combine these terms together to yield a matrix, we first construct the non-diagonal terms, which we obtained from solving the integral in the case where $i \neq j$.
\begin{equation}
\phi(z_i^+)\phi(z_j^+)\prod_{k \neq i, j}\Phi(-z_k^+) = \frac{\phi(z_i^+)\phi(z_j^+)}{\Phi(-z_i^+)\Phi(-z_j^+)}\prod_k\Phi(-z_k^+)
\end{equation}
Upon inspection, the outer product of $\vw$ will yield the correct non-diagonal elements of said matrix
\begin{equation}
    \vw\vw^\top \prod_k\Phi(-z_k^+)
\end{equation}
We note that this will result in diagonal terms where the $ii^{th}$ element corresponds to 
\begin{equation}
\frac{\phi(z_i^+)^2}{\Phi(-z_i^+)^2}\prod_k\Phi(-z_k^+)
\end{equation}
This can be expressed as
\begin{equation}
    \vw \odot \vw \prod_k\Phi(-z_k^+)
\end{equation}
and will need to be corrected when building the matrix. Now, we proceed with the diagonal terms. We return to the case where $i = j$ and rewrite the results by expanding the terms
\begin{subequations}
    \begin{align}
    \left(z_i^+\,\phi(z_i^+) + \Phi(-z_i^+)\right)\prod_{i \neq k} \Phi(-z_k^+) &= z_i^+\,\phi(z_i^+) \prod_{i \neq k} \Phi(-z_k^+) + \prod_{k} \Phi(-z_k^+)\\
    &= \left(1 + \frac{z_i^+\,\phi(z_i^+)}{\Phi(-z_i^+)}\right)\prod_{k} \Phi(-z_k^+)
    \end{align}
\end{subequations}
Therefore, the corrected diagonal can be expressed as
\begin{equation}
    (\mathbf{1} + \vz^+ \odot \vw - \vw \odot \vw)\prod_{k} \Phi(-z_k^+)
\end{equation}
Combining the diagonal and non-diagonal terms into a matrix will yield
\begin{equation}
    \left(\vw \vw^\top + \text{diag}\left(\mathbf{1} + (\vz^+ - \vw) \odot \vw \right)\right)\prod_i \Phi(-z_i^+)
\end{equation}
Recall that \eqref{eq: three_terms_3} requires the evaluation of
\begin{equation}
    \sum_{i,j=1}^d \Sigmanabla_{ij}I_{ij}
\end{equation}
The element-wise multiplication and summation is the definition of the Frobenius inner product, which can also be expressed as the trace of the matrix products. Hence, \eqref{eq: three_terms_3} becomes
\begin{equation}
    \text{tr}\bigg(\Sigmanabla\left(\vw \vw^\top + \text{diag}\left(\mathbf{1} + (\vz^+ - \vw) \odot \vw \right)\right)\bigg)\prod_{i = 1}^d \Phi(-z_i^+)
\end{equation}

\eqref{eq: three_terms} is equivalent to
\begin{subequations}
\label{eq: three_terms_simplified}
    \begin{align}
    \label{eq: three_terms_simplified_1} 
        &=\bigg(\prod_{i = 1}^d \Phi(-z_i^+)\bigg)\bigg(\munabla^\top \munabla - \|\nabla f(\vx^+)\|_2^2 + 2\munabla^\top \cholesky \mathbf{w}\\
    \label{eq: three_terms_simplified_2}
        &+\text{tr}\big(\Sigmanabla\left(\vw \vw^\top + \text{diag}\left(\mathbf{1} + (\vz^+ - \vw) \odot \vw \right)\right)\big)
        \bigg)
    \end{align}
\end{subequations}

The resulting optimizer of the acquisition function can be obtained from substituting the terms from \eqref{eq: EI_g_lower_bound}
\begin{subequations}
\label{eq: final_AF}
    \begin{align}
    \label{eq: final_AF_1}
        \vx_{t+1} %&= \underset{\vx}{\arg\max}\ \text{EI}_{\f} - \alpha\text{EI}_{\nablaf}\\
        &\leftarrow \underset{\vx}{\arg\max}\ (\mu(\vx) - f(\vx^+))\Phi\left(\frac{\mu(\vx) - f(\vx^+)}{\sigma(\vx)}\right) + \sigma(\vx)\phi\left(\frac{\mu(\vx) - f(\vx^+)}{\sigma(\vx)}\right)\\
    \label{eq: final_AF_2}
        &-\alpha\bigg(\prod_{i = 1}^d \Phi(-z_i^+)\bigg)\bigg(\munabla^\top \munabla - \|\nabla f(\vx^+)\|_2^2 + 2\munabla^\top \cholesky \mathbf{w}\\
    \label{eq: final_AF_3}
        &+\text{tr}\big(\Sigmanabla\left(\vw \vw^\top + \text{diag}\left(\mathbf{1} + (\vz^+ - \vw) \odot \vw \right)\right)\big)
        \bigg)
    \end{align}
\end{subequations}

\subsection{Computational Complexity of EI-GN}
\label{app: complexity}
Recall that EI-GN is comprised of two terms EI$_f$ and $\overline{\text{EI}}_s$. Since the former corresponds to canonical EI, the complexity of this AF is dominated by $\overline{\text{EI}}_s$. Inspection of its closed-form expression in \eqref{eq: three_terms_simplified} shows it has complexity $O(d^2)$ due to matrix--vector products (e.g., $\munabla^\top \mathbf{L}\vw$) and quadratic forms/traces involving $\Sigmanabla$. 

However, since our implementation utilizes independent GPs that lead to diagonal posterior covariance $\Sigmanabla$ (and hence diagonal $\mathbf{L}$). Under this diagonal structure, all terms decompose into elementwise sums (e.g., $\munabla^\top \mathbf{L}\vw=\sum_i \mu^\nabla_{i}(\vx) (\cholesky)_{ii} w_i$ and $\mathrm{tr}(\Sigmanabla \vw\vw^\top)=\sum_i (\sigma_{i}^{\nabla}(\vx))^2 w_i^2$), so acquisition evaluation costs $O(d)$ per candidate $\vx$, in addition to the cost of GP posterior evaluation.

In either case, the acquisition evaluation cost is typically negligible relative to GP posterior evaluation, which requires kernel evaluations and linear solves (e.g., Cholesky-based solves) whose cost is $\mathcal{O}(N^3)$. In the pessimistic scenario where the independent GPs cannot be trained or inferred in parallel, the cost can be up to $\mathcal{O}((d+1)N^3)$, which still results in negligible cost for EI-GN evaluation. 

% \subsection{Gradient-Enhanced Gaussian Processes}
% As stated in Section~\ref{sec: model}, independent GPs for $f$ and each of the dimensions in $\nabla f$

% The choice of parallel models reduces complexity to only $\mathcal{O}((d+1)N^3)$. Assuming these additional models are trained in parallel, this further reduces complexity to $\mathcal{O}(N^3)$, independent of $d$. We model each gradient with similar equations to \eqref{eq: GP_mean} and \eqref{eq: GP_var}, except each dimension has their corresponding mean and kernel functions. We shall denote the posterior gradient mean and covariance as $\mu^\nabla$ and $\Sigma^\nabla$, respectively. Direct comparisons to gradient-enhanced surrogate models are beyond the scope of this work, as they modify the posterior and will confound our focus on acquisition design.

\section{Implementation Details}
\label{app: implementation}
\subsection{Gaussian Processes}
Our GP surrogates use the GPyTorch implementation~\cite{gardner2018gpytorch}. Unless otherwise noted, we use a Mat\'ern-$5/2$ kernel with ARD, with lengthscale hyperpriors $\mathrm{LogNormal}(\log 0.4,\,0.7)$ and outputscale hyperpriors $\mathrm{Gamma}(2,\,0.5)$. For within-model comparisons, we use an RBF kernel to match the GP prior used to sample the objective. Inputs are normalized to $[0,1]^d$ and outputs are standardized to zero mean and unit variance. 

To model gradient information, we fit $d$ independent GP surrogates to the observed partial derivatives, one per coordinate of $\nabla y$, using the same preprocessing and training procedure as for $y$. Each gradient GP has its own hyperparameters, fit by marginal likelihood, and provides posterior moments for $\nabla f(\vx)$ used in EI-GN. This choice treats $f$ and $\nabla f$ as separate outputs and does not enforce the exact joint GP consistency constraints that arise from differentiating a single GP over $f$; we adopt it for simplicity and scalability. In practice, gradient observations can span a large range of magnitudes, which can lead to occasional numerical issues in GP training. We therefore use an adaptive Cholesky jitter, increasing from $10^{-9}$ up to $10^{-2}$ only if factorization fails.

\subsection{Acquisition Function Optimization}
We implement EI-GN in BoTorch~\cite{balandat2020botorch} and maximize the acquisition using multi-start L-BFGS. We first generate a set of candidate points via Boltzmann sampling over the acquisition (raw samples), then select the top-$k$ candidates as starting points (num restarts) for optimization.

\subsection{Synthetic Benchmarks}
\begin{table}[h]
\caption{Synthetic benchmark implementation details for Shekel (4$d$), Hartmann (6$d$), Cosine (8$d$), Griewank (10$d$), and Ackley (14$d$). For functions defined as minimization problems, we negate the objective for maximization.}
\centering
\begin{tabular}{l c c c c c c}
\toprule
Function & $d$ & Initial points & Budget & Input space & Raw samples & Num restarts \\
\midrule
Shekel & 4 & 12 & 125 & $[0, 10]^4$ & 8 & 256 \\
Hartmann & 6 & 18 & 150 & $[0, 1]^6$ & 10 & 512 \\
Cosine & 8 & 24 & 200 & $[-1, 1]^8$ & 20 & 1024 \\
Griewank & 10 & 30 & 300 & $[-10, 10]^{10}$ & 20 & 1024 \\
Ackley & 14 & 42 & 500 & $[-5, 5]^{14}$ & 20 & 1024 \\
\bottomrule
\end{tabular}
\label{tab: synthetic_table}
\end{table}
We perform experiments across 20 seeds and report mean $\pm$ one standard error for Shekel (4$d$), Hartmann (6$d$), Cosine (8$d$), Griewank (10$d$), and Ackley (14$d$) \cite{surjanovic2014virtual}. Table~\ref{tab: synthetic_table} shows the implementation details for each of the synthetic benchmarks.

\subsection{GP Samples}

\begin{table}[h]
\caption{Implementation details for within and out of model comparisons in 7$d$, 8$d$, and 9$d$.}
\centering
\begin{tabular}{l c c c c c c}
\toprule
Function & $d$ & Initial points & Budget & Kernel & Raw samples & Num restarts \\
\midrule
Within & 7 & 21 & 175 & RBF & 10 & 512 \\
Within & 8 & 24 & 200 & RBF & 20 & 1024 \\
Within & 9 & 27 & 250 & RBF & 20 & 1024 \\
Out & 7 & 21 & 175 & Mat\'ern-$5/2$ & 10 & 512 \\
Out & 8 & 24 & 200 & Mat\'ern-$5/2$ & 20 & 1024 \\
Out & 9 & 27 & 250 & Mat\'ern-$5/2$ & 20 & 1024 \\
\bottomrule
\end{tabular}
\label{tab: gp_samples_table}
\end{table}

We use the objectives for within \cite{hennig2012entropy} and out of model comparisons \cite{muller2021local}, but create a separate one for each $d$. First, we create GP priors with RBF kernels, zero means, and lengthscales $0.4/\sqrt{d}$. $200d$ Sobol points in $[0, 1]^d$ are jointly sampled and GPs with the same hyperparameters are fit accordingly; the posterior means are treated as the black-box objective $f$. To maintain correct model specifications, we do not fit the surrogate between BO iterations for the within model comparisons, and also use the same hyperparameters as the GP prior. As for the out of model comparisons, we revert to the Mat\'ern-$5/2$ kernel with ARD, with lengthscale hyperpriors $\mathrm{LogNormal}(\log 0.4,\,0.7)$ and outputscale hyperpriors $\mathrm{Gamma}(2,\,0.5)$. Table~\ref{tab: gp_samples_table} summarizes the within and out of model comparison implementation details.

\subsection{Policy Search}
We use stochastic policies and estimate gradients with the policy gradient theorem. To reduce estimated variance, we subtract a baseline given by the rollout-average return.
\[
\nabla_\theta J \approx \frac{1}{M} \sum_{i=1}^{M} (R_i - \bar{R}) \nabla_\theta \log \pi_\theta(\tau_i),
\]
where $\bar{R} = \frac{1}{N}\sum_i R_i$ is the mean return across $M$ rollouts.

\subsubsection{Acrobot}
Acrobot \cite{sutton1995generalization} is a two-link pendulum swing-up task. We modify the reward to an via energy-based shaping
\begin{equation}
    r = -0.001\left(\cos(\theta_1) + \cos(\theta_1 + \theta_2)\right),\notag
\end{equation}
which provides a dense signal that correlates with the height of the end-effector and stabilizes policy-search optimization compared to sparse, goal-only rewards. This shaping preserves the swing-up objective (maximized when the links are upright) while yielding a smoother return landscape for global optimization methods.

The observation space is 6-dimensional: $[\cos\theta_1, \sin\theta_1, \cos\theta_2, \sin\theta_2, \dot\theta_1, \dot\theta_2]$, while we use a bounded linear policy: $a = \tanh(\mathbf{o}^\top \mathbf{w})$, where $\mathbf{w} \in \mathbb{R}^6$, yielding a 6-dimensional search space. Episodes run for 500 steps with 32 Monte Carlo rollouts for gradient estimation. The policy noise standard deviation is $\sigma_\pi = 0.05$. Parameter bounds are $[-1, 1]^6$.
We initialize with 18 Sobol points and set a budget of 100 episodes. For EI-GN and EI, we use 32 restarts and 512 raw samples for AF optimization.

\subsubsection{CartPole}
CartPole \cite{barto1988neuronlike} is a control balancing task with a 4-dimensional observation: $[x, \dot{x}, \theta, \dot\theta]$ (cart position, cart velocity, pole angle, angular velocity). We replace the sparse survival reward with a quadratic shaping cost around the upright equilibrium,
\begin{equation}
r = -\left(\theta_t^{2} + 0.1\,\dot{\theta}_t^{2} + 0.01\,x_t^{2} + 0.001\,\dot{x}_t^{2} + 0.001\,a_t^{2}\right),\notag
\end{equation}
which yields a smooth, dense return signal for continuous policy optimization. Termination conditions and dynamics follow the standard CartPole balancing setup; the modification affects only the reward signal used to define the policy-search objective. We use a 1-hidden-layer neural policy with 2 hidden units and ReLU activation:
\[
a = \tanh(\text{ReLU}(\mathbf{o}^\top \mathbf{w}_1 + \mathbf{b}_1)^\top \mathbf{w}_2 + \mathbf{b}_2),
\]
where $\mathbf{w}_1 \in \mathbb{R}^{4 \times 2}$, $\mathbf{b}_1 \in \mathbb{R}^2$, $\mathbf{w}_2 \in \mathbb{R}^2$, and $\mathbf{b}_2 \in \mathbb{R}$. This yields a 13-dimensional search space. Episodes run for 200 steps with 48 rollouts. Parameter bounds are $[-2, 2]^{13}$. 39 initial Sobol points are used and it is run for 100 episodes. AF optimization is done via 32 restarts and 512 raw samples.

% \subsection{Numerical Conditioning}
% \label{app: EIGN normalization}
% During acquisition maximization, we normalize the objective and stationarity components over the optimizer’s temporary candidate pool at each BO iteration to balance their magnitudes and improve numerical stability. This induces an iteration-dependent effective trade-off between the two terms within the inner optimization.
% During acquisition maximization, we apply an affine normalization to the two EI-GN terms over the optimizer’s temporary candidate pool at each BO iteration (i.e., the sampled candidates evaluated during acquisition maximization). This is solely to improve numerical conditioning of the inner optimization and does not alter the EI-GN definition—only the numerical scale on which the optimizer operates.

\color{black}
\section{Additional Experiments}
\label{app: ablation}
\subsection{Separable Truncation and $\overline{\text{EI}}_s$}
\label{sec: orthant_approximation}
To better comprehend $\overline{\text{EI}}_s$ \eqref{eq: EI_s overline}, we perform additional experiments comparing EI-GN (based on the tractable form $\overline{\text{EI}}_s$) against EI$_g$ Lower Bound (based on the original norm-based truncation). Since EI$_g$ Lower Bound is intractable, we estimate it via Monte Carlo using the reparameterization trick.

\begin{figure}[tb]
\centering
\includegraphics[width=0.48\textwidth]{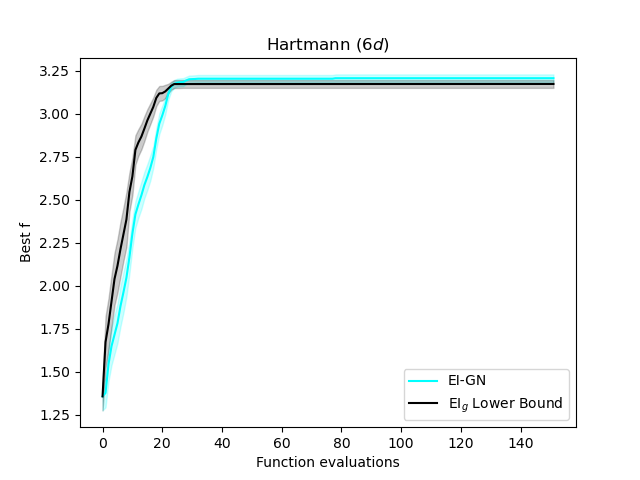}
\includegraphics[width=0.48\textwidth]{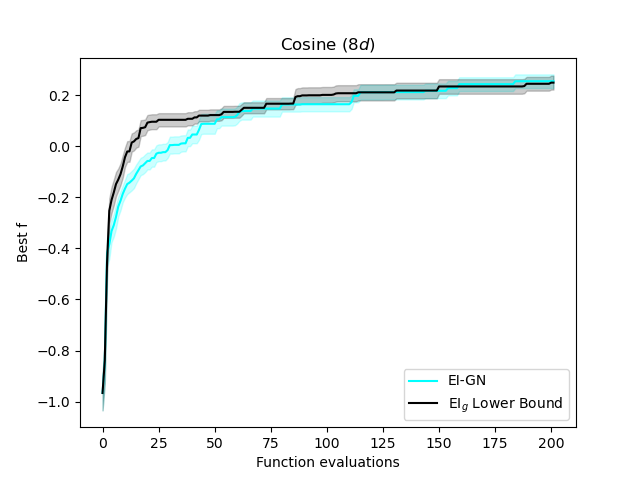}
\caption{Results displaying mean $\pm$ one standard error for Hartmann (6$d$), Cosine (8$d$) for EI-GN, EI$g$ Lower Bound.}
\label{fig: orthant approximation}
\end{figure}

% \begin{table}[h]
% \caption{Best $f$ (mean $\pm$ standard error across 20 seeds) found for Hartmann (6$d$) for EI-GN and EI$_g$ Lower Bound.}
% \centering
% \begin{tabular}{lccc}
% \toprule
% Method & Eval 50 & Eval 100 & Eval 150 \\
% \midrule
% EI-GN & 3.20 $\pm$ 0.02 & 3.20 $\pm$ 0.02 & 3.20 $\pm$ 0.02 \\
% EI$_g$ Lower Bound & 3.18 $\pm$ 0.03 & 3.19 $\pm$ 0.02 & 3.19 $\pm$ 0.02 \\
% \bottomrule
% \end{tabular}
% \label{tab:hartmann_orthant}
% \end{table}

% \begin{table}[h]
% \caption{Best $f$ (mean $\pm$ standard error across 20 seeds) found for Cosine (8$d$) for EI-GN and EI$_g$ Lower Bound.}
% \centering
% \begin{tabular}{lcccc}
% \toprule
% Method & Eval 50 & Eval 100 & Eval 150 & Eval 200 \\
% \midrule
% EI-GN & 0.11 $\pm$ 0.05 & 0.18 $\pm$ 0.04 & 0.23 $\pm$ 0.03 & 0.26 $\pm$ 0.03 \\
% EI$_g$ Lower Bound & 0.14 $\pm$ 0.04 & 0.20 $\pm$ 0.03 & 0.24 $\pm$ 0.03 & 0.26 $\pm$ 0.03 \\
% \bottomrule
% \end{tabular}
% \label{tab:cosine_orthant}
% \end{table}

Figure~\ref{fig: orthant approximation} displays the performance of EI-GN and EI$_g$ Lower Bound on Hartmann (6$d$) and Cosine (8$d$) for mean and one standard error of best $f$ across 20 seeds. In Hartmann (6$d$), the results indicate virtually identical performance since the difference in the means are within standard error. The same can be said for Cosine (8$d$), but EI$_g$ Lower Bound has marginally stronger performance in the early iterations. These results indicate that the mean-field approximation $\overline{\text{EI}}_s$ used in EI-GN yields similar empirical behavior to the EI$_g$ Lower Bound.

% (i) gradient-enhanced EI using joint GPs (dEI), and (ii) the combined approach (dEI-GN), which integrates EI-GN with gradient-enhanced posteriors. This enables a controlled comparison that separates gains due to improved posterior modeling from those due to acquisition design.

% \begin{table}[h]
% \caption{Best $f$ (mean $\pm$ std error across 20 seeds) for Holder Table (2$d$).}
% \centering
% \begin{tabular}{lcc}
% \toprule
% Method & Eval 25 & Eval 50 \\
% \midrule
% \textbf{dEI-GN} & \textbf{18.2 $\pm$ 0.6} & \textbf{19.1 $\pm$ 0.3}\\
% dEI & 16.3 $\pm$ 0.5 & 16.8 $\pm$ 0.4\\
% EI-GN & 16.5 $\pm$ 0.6 & 17.1 $\pm$ 0.5\\
% EI & 13.0 $\pm$ 1.1 & 13.5 $\pm$ 0.9\\
% \bottomrule
% \end{tabular}
% \end{table}

% \begin{table}[h]
% \caption{Best $f$ (mean $\pm$ std error across 20 seeds) for Shekel (4$d$).}
% \centering
% \begin{tabular}{lccc}
% \toprule
% Method & Eval 40 & Eval 80 & Eval 125\\
% \midrule
% \textbf{dEI-GN} & \textbf{4.98 $\pm$ 0.52} & \textbf{5.01 $\pm$ 0.58} & \textbf{5.03 $\pm$ 0.57}\\
% dEI & 4.18 $\pm$ 0.47 & 4.42 $\pm$ 0.46 & 4.53 $\pm$ 0.48\\
% EI-GN & 3.05 $\pm$ 0.38 & 3.88 $\pm$ 0.47 & 4.03 $\pm$ 0.42\\
% EI & 0.82 $\pm$ 0.09 & 0.93 $\pm$ 0.08 & 1.02 $\pm$ 0.08\\
% \bottomrule
% \end{tabular}
% \end{table}

\color{black}

% \begin{table}[h]
% \centering
% \begin{tabular}{l c c c c c c}
% \toprule
% Function & $d$ & Initial points & Budget & Raw samples & Num restarts \\
% \midrule
% Shekel & 4 & 12 & 125 & 8 & 256 \\
% Hartmann & 6 & 18 & 150 & 10 & 512 \\
% Cosine & 8 & 24 & 200 & 20 & 1024 \\
% Griewank & 10 & 30 & 300 & 20 & 1024 \\
% Ackley & 14 & 42 & 500 & 20 & 1024 \\
% \bottomrule
% \end{tabular}
% \caption{Synthetic benchmark implementation details for Shekel (4$d$), Hartmann (6$d$), Cosine (8$d$), Griewank (10$d$), and Ackley (14$d$). For functions defined as minimization problems, we negate the objective for maximization.}
% \label{tab: synthetic_table}
% \end{table}
\color{black}
\subsection{Hyperparamter tuning for $\alpha$}
\label{app: hyperparameter alpha}
\begin{figure*}[tb]
    \centering
    \includegraphics[width=\textwidth]{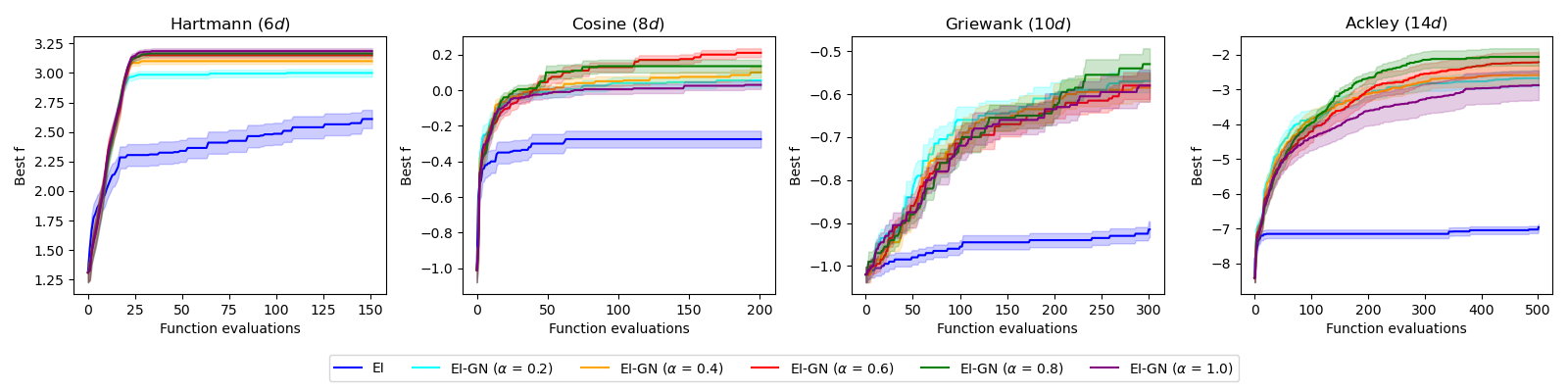}
    \caption{Hyperparameter tuning of $\alpha \in [0,1]$ displayed with mean $\pm$ one standard error for synthetic benchmarks: Hartmann (6$d$), Cosine (8$d$), Griewank (10$d$), and Ackley (14$d$). EI-GN with $\alpha = 0$ is displayed as EI for brevity.}
    \label{fig: synthetic_ablation}
\end{figure*}
In implementation, the objective-improvement and stationarity components can differ substantially in scale across problems and across BO iterations. During acquisition maximization, we apply an online rescaling of the objective-improvement and stationarity components using mean and standard deviation computed over the optimizer’s temporary candidate pool at each BO iteration to keep magnitudes comparable and improve numerical stability. The scaling does not eliminate the role of $\alpha$, but makes a consistent sweep over $\alpha \in [0,1]$ meaningful across tasks. We focus on this range since $\alpha=0$ recovers EI and larger values place progressively greater emphasis on the stationarity component within the normalized acquisition; we did not find additional benefit outside this range in our tuning sweeps.

% \label{app: ablation_synthetic}
% \begin{figure*}[tb]
%     \centering
%     \includegraphics[width=\textwidth]{Figures/within_ablation.png}
%     \caption{Hyperparameter tuning of $\alpha \in [0,1]$ displayed with mean $\pm$ one standard error for within-model comparisons in 7$d$, 8$d$, and 9$d$ for EI-GN with varying $\alpha \in [0,1]$. EI-GN with $\alpha = 0$ is displayed as EI for brevity.}
%     \label{fig: within model_ablation}
% \end{figure*}
% \begin{figure*}[tb]
%     \centering
%     \includegraphics[width=\textwidth]{Figures/out of model_ablation.png}
%     \caption{Hyperparameter tuning of $\alpha \in [0,1]$ displayed with mean $\pm$ one standard error for out-of-model comparisons in 7$d$, 8$d$, and 9$d$ for EI-GN with varying $\alpha \in [0,1]$. EI-GN with $\alpha = 0$ is displayed as EI for brevity.}
%     \label{fig: out of model_ablation}
% \end{figure*}

Figure~\ref{fig: synthetic_ablation} show EI-GN with different $\alpha$; when $\alpha = 0$, the lack of gradient information  leads to over-exploitation and subpar performance, as expected of EI. All other trajectories with positive $\alpha$ generally achieve similar performance, suggesting the weight of the hyperparameter is largely insignificant. % However, EI-GN ($\alpha = 0.2$) for out of model (9$d$) comparisons in Figure~\ref{fig: out of model_ablation}  underperforms compared to $\alpha \geq 0.4$. Out of all the GP sample-based benchmarks, this problem has the most challenging objective landscape and optimization conditions, indicating a greater weight on the gradient-term is necessary.

\end{document}